\pdfoutput=1

\documentclass[11pt]{article}

\usepackage{authblk}

\usepackage[final]{ACL2023}

\usepackage{times}
\usepackage{latexsym}

\usepackage[T1]{fontenc}

\usepackage[utf8]{inputenc}

\usepackage{microtype}

\usepackage{inconsolata}

\usepackage{subfigure}
\usepackage{tabularx}
\usepackage{booktabs}
\usepackage{tikz}
\usepackage{pgfplots}
\usepackage{amsmath}
\usepackage{amssymb}
\usepackage{graphicx}
\usepackage{multirow}
\usepackage{subcaption}
\usepackage{spverbatim}
\usepackage{xcolor}         
\usepackage{colortbl}
\usepackage{color}
\usepackage{makecell}
\usepackage{float}

\usepackage{times}
\usepackage{latexsym}
\usepackage{caption}
\usepackage{mismath}
\usepackage{tcolorbox}
\usepackage{array}
\usepackage{pifont}
\usepackage{tabularx}
\usepackage{adjustbox}
\usepackage{enumitem}
\usepackage{xspace}
\usepackage{amsfonts}
\usepackage{hyperref}
\usepackage{url}
\usepackage{makecell}

\definecolor{Gray}{gray}{0.90}
\definecolor{lightgrey}{gray}{0.925}

\definecolor{lightcyan}{RGB}{178, 255, 255}
\definecolor{lightpurple}{RGB}{203, 195, 227}
\definecolor{lightgreen}{RGB}{224,255,210}

\definecolor{lightyellow}{RGB}{255,255,224}

\definecolor{mypink1}{RGB}{255, 204, 204}
\definecolor{mygrey1}{RGB}{204,229,255}
\definecolor{myblue1}{RGB}{200, 226, 255}
\definecolor{mycyan1}{RGB}{0, 255, 255}
\definecolor{mygreen1}{RGB}{182,215,168}
\definecolor{myyellow1}{RGB}{225,165,0}
\definecolor{myorange1}{RGB}{255,215,0}

\definecolor{hookersgreen}{rgb}{0.0, 0.44, 0.0}
\definecolor{indiagreen}{rgb}{0.07, 0.53, 0.03}
\definecolor{islamicgreen}{rgb}{0.0, 0.56, 0.0}
\definecolor{kellygreen}{rgb}{0.3, 0.73, 0.09}
\definecolor{alizarin}{rgb}{0.82, 0.1, 0.26}
\usepackage[normalem]{ulem}
\usepackage{pifont}
\newcommand{\cmark}{{\color{kellygreen} \ding{51}}}
\newcommand{\xmark}{{\color{alizarin} \ding{55}}}

\title{Large Language Models as Zero-shot \\Dialogue State Tracker through Function Calling}

\usepackage{authblk}

\author[1]{\bf Zekun Li\textsuperscript{$\dagger$}}
\author[2]{\bf Zhiyu Zoey Chen}
\author[3]{\bf Mike Ross}
\author[3]{\bf Patrick Huber}
\author[3]{\bf Seungwhan Moon}
\author[3]{\\ \bf Zhaojiang Lin}
\author[3]{\bf Luna Dong}
\author[3]{\bf Adithya Sagar}
\author[1]{\bf Xifeng Yan}
\author[3]{\bf Paul A. Crook}

\affil[1]{University of California, Santa Barbara}
\affil[2]{Carnegie Mellon University}
\affil[3]{Meta AI}

\begin{document}
\maketitle
\begin{abstract}
Large language models (LLMs) are increasingly prevalent in conversational systems due to their advanced understanding and generative capabilities in general contexts. 
However, their effectiveness in task-oriented dialogues (TOD), which requires not only response generation but also effective dialogue state tracking (DST) within specific tasks and domains, remains less satisfying.  
In this work, we propose a novel approach \textsc{FnCTOD} for solving DST with LLMs through function calling. This method improves zero-shot DST, allowing adaptation to diverse domains without extensive data collection or model tuning. 
Our experimental results demonstrate that our approach achieves exceptional performance with both modestly sized open-source and also proprietary LLMs: with in-context prompting it enables various 7B or 13B parameter models to surpass the previous state-of-the-art (SOTA) achieved by ChatGPT, and improves ChatGPT's performance beating the SOTA by 5.6\% average joint goal accuracy (JGA). Individual model results for GPT-3.5 and GPT-4 are boosted by 4.8\% and 14\%, respectively.
We also show that by fine-tuning on a small collection of diverse task-oriented dialogues, we can equip modestly sized models, specifically a 13B parameter LLaMA2-Chat model, with function-calling capabilities and DST performance comparable to ChatGPT while maintaining their chat capabilities. We have made the code publicly available.\footnote{\url{https://github.com/facebookresearch/FnCTOD}}



\end{abstract}

{\let\thefootnote\relax\footnotetext{\hspace{-1.5mm}$\dagger$\hspace{0.2mm}Work undertaken while interning at Meta.}}

{\let\thefootnote\relax\footnotetext{\hspace{-1.5mm}*\hspace{0.2mm}Correspondence authors: \texttt{zekunli@ucsb.cs.edu}, and \texttt{pacrook@meta.com}.}}

\section{Introduction}
Recent years have seen the rapid development of large language models (LLMs) that have demonstrated exceptional natural language understanding and generation capabilities. The integration of LLMs into industry applications, particularly as conversational assistants, is a notable trend. 
Fine-tuned with conversations between users and assistants, these models are further aligned with human preferences to enhance their ability to deliver fluent, helpful, and polite responses to user inquiries.
Notable examples include proprietary systems such as ChatGPT\footnote{http://chatgpt.openai.com/} and Claude\footnote{https://www.anthropic.com/index/introducing-claude}, as well as open-source models such as LLaMA2-Chat~\cite{llama2}, Vicuna~\cite{vicuna2023}, Baichuan~\cite{baichuan2}.

\begin{figure}[!t]
\begin{center}
\includegraphics[width=1\linewidth]{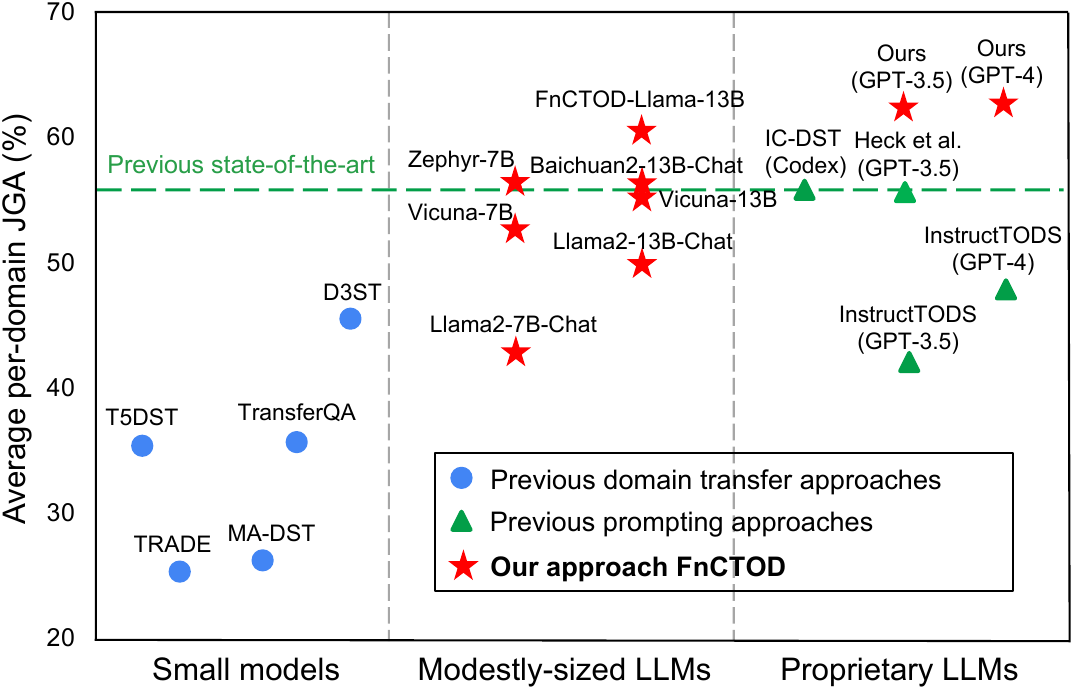}
\end{center}
   \caption{\textbf{Zero-shot DST performance comparison} among (1) previous domain transfer approaches using small models; (2) previous prompting approaches exclusively relying on advanced proprietary LLMs; and (3) our approach, compatible with various LLMs, empowers various 7B and 13B models for superior performance and sets new state-of-the-art with GPT-4.
   }\label{fig:crown-jewel}
\end{figure}

\begin{table*}[t!]
\caption{
\textbf{Comparison of different zero-shot DST paradigms.} Plug-\&-Play means the (chat-tuned) LLMs can be
equipped with this capability, preserving their conversational capabilities.
}
\centering
\setlength{\tabcolsep}{3pt}
\scriptsize
\begin{tabular}{@{}l c c c c@{}}
\toprule
{\bf Zero-shot DST Paradigms} & {\bf Base Model} & {\bf Fine-tuning} &  {\bf Prompting} &
{\bf Plug-\&-Play}
\\ \midrule
Domain transfer approaches~\citep{lin-etal-2021-zero,lin-etal-2021-leveraging,D3ST} & Small LMs  & \cmark   & \xmark  & \xmark   \\ \midrule
Previous prompting approaches~\cite{heck2023chatgpt,instructtods}  & Advanced proprietary LLMs  &  \xmark  &   \cmark  &  \xmark  \\ \midrule
\textsc{FnCTOD} ({\bf Ours}) & \makecell{{Modestly-sized open-source LLMs}\\{\& Advanced proprietary LLMs}}  &  \cmark  &   \cmark  &  \cmark \\
\bottomrule
\end{tabular}
\label{tab:motivation}
\end{table*}

The primary focus of these chat-tuned LLMs has typically been on responding in general contexts. However, for another important type of conversation, task-oriented dialogues (TOD), the model is required to extract the intentions of users at each turn of the conversation, represented as slot-value pairs of per-domain predefined schemas; a process known as Dialogue State Tracking (DST). 
The challenge lies in the model's ability to accurately summarize user needs over multiple turns of conversation and also
strictly adhere to a domain-specific ontology.
The most direct solutions~\cite{simpletod,soloist,pptod} necessitate training on curated domain-specific annotated data, a process that is notoriously costly and labor-intensive. Despite efforts in data augmentation~\cite{li2020coco} and automated dataset creation using GPT-3~\cite{li2022controllable}, these methods struggle to generalize to unseen domains. 
To achieve zero-shot DST for unseen domains, prior approaches usually involved domain transfer methods~\citep{campagna2020zero,transferqa,D3ST}. While such approaches are trained on alternative domains, they still require domains with matching annotation schema, and their performance has been far from satisfactory.

LLMs exhibit remarkable capabilities for tackling various tasks without the need for task-specific fine-tuning, making them suited for zero-shot DST. However, while there have been initiatives to leverage ChatGPT for zero-shot DST~\cite{icdst,hudevcek2023llms,heck2023chatgpt,instructtods}, these methods tend to treat DST as a standalone task rather than chat completion, which the models, especially chat-tuned models, are more proficient in.
They usually take the whole conversation as input along with detailed instructions to generate in domain-specific formats. This setup poses challenges due to the long task context and specific output requirements. Consequently, this works exclusively with advanced ChatGPT or Codex models but fails with less powerful LLMs~\cite{hudevcek2023llms}.

In this work, we introduce a novel approach \textsc{FnCTOD}, to address zero-shot DST with LLMs. Our method seamlessly integrates DST as a part of the assistant's output during chat completion. Specifically, we treat the schema of each task-oriented dialogue domain as a specific function, and DST for this domain as the process of ``calling'' the corresponding function. We thus instruct LLMs to generate function calls along with the response in the assistant's output. To achieve this, we convert the domain schema into function specifications, which include the function's description and required arguments, and incorporate them into the \emph{system prompt} of the LLM. Additionally, we integrate these function calls into the assistant's output within the \textit{dialogue context}.

As shown in Figure~\ref{fig:crown-jewel}, experimental results on the MultiWOZ benchmark~\cite{budzianowski-etal-2018-multiwoz} represent a significant milestone. Our approach is the first that, without further fine-tuning, enables modestly sized open-source LLMs (7B or 13B parameters) to achieve comparable or superior performance compared to previous state-of-the-art (SOTA) prompting methods that relied exclusively on advanced proprietary LLMs such as ChatGPT and Codex~\cite{hudevcek2023llms,heck2023chatgpt,instructtods}.
Furthermore, our approach beats the previous zero-shot SOTA by 5.6\% Av. JGA, firmly establishing a new standard. It improves ChatGPT performance; beating previous individual best results for GPT-3.5 and GPT-4 by 4.8\% and 14\%, respectively.

Additionally, we show that by fine-tuning a 13B \textsc{LLaMA2-Chat} model using a collection of 7,200 task-oriented dialogues --- consisting of 200 randomly selected dialogues covering 36 diverse domains, from heterogeneous TOD datasets --- we can equip
it with function-calling DST abilities comparable to ChatGPT while still maintaining its response generation capabilities.

The comparison with prior studies
is summarized in Table~\ref{tab:motivation} and Figure~\ref{fig:crown-jewel}. {\bf Our contribution is threefold:}
\textbf{(1)} Demonstration that the FnCTOD approach achieves outstanding performance with both open-source and proprietary LLMs through \emph{in-context prompting}: enables open-source 7--13B models to surpass the previous SOTA achieved by ChatGPT, and enhances GPT-4's performance by 14\%, establishing a new SOTA. 
\textbf{(2)} Bridging the \emph{zero-shot} DST performance gap between open-source models and ChatGPT by fine-tuning on a small collection of diverse dialogues.
\textbf{(3)} Showing that function calling DST capabilities can be integrated into existing chat-tuned LLMs while preserving response capabilities.


\section{Related Work}
\subsection{Dialogue State Tracking}
DST is an essential, yet challenging task in the construction of TOD systems. Its primary purpose is to extract and track user goals at each turn throughout the conversation. The tracked dialogue state is usually represented in the slot values of the predefined schema for specific domains. This requires the slot values to adhere closely to the domain-specific schema.
Consequently, previous methods have relied on the collection and annotation of domain-specific dialogues for model training~\cite{lee2019sumbt,trade,heck2020trippy,simpletod,soloist,lin2020mintl}
However, obtaining training data is notoriously expensive, even with data augmentation methods~\cite{li2020coco} and methods that utilize GPT-3 to automatically simulate such data~\cite{li2022controllable}. Furthermore, these approaches are limited to handling only the domains covered in the training data.

To address zero-shot DST in unseen domains, previous cross-domain transfer strategies based on small models typically leverage extra dialogue corpora in similar domains~\cite{wu-etal-2020-improving-limited,lin-etal-2021-zero,pptod} or redefine DST in terms of alternative tasks, e.g., Q\&A~\cite{lin-etal-2021-leveraging} or summarization~\cite{shin-etal-2022-dialogue} to find additional data. Despite these efforts, their overall zero-shot performance remains relatively low.

\subsection{Leveraging LLMs for Dialogue Tasks}
LLMs~\cite{gpt3,palm,gpt4} have demonstrated remarkable capabilities in handling various tasks without further fine-tuning. Recent chat/instruction-tuned models further exhibit impressive performance in conversational contexts~\cite{llama2,vicuna2023,yang2023baichuan}. However, current chat models primarily focus on general conversation, often omitting task-oriented dialogues (TOD).
TOD differs from general conversation in that it requires models to not only generate responses but also track dialogue states according to domain-specific schemas. 
While ChatGPT has shown effectiveness in response generation for TOD tasks~\cite{li2024guiding}, its performance in zero-shot DST, as explored in recent research on prompting approaches~\cite{icdst,bang2023multitask,hudevcek2023llms,heck2023chatgpt,zhang2023sgp,instructtods}, is unsatisfactory and therefore remains as a open challenge.



\subsection{Tool Use within LLMs}
Early work on tool use ~\cite{talm,toolformer} and the recent launch of GPT-4 plug-in and function calling features~\cite{gpt4}, have highlighted the importance of function calling for LLMs, encouraging follow-up work~\cite{gorilla,hugginggpt,apibank}. Commonly integrated tools include web browsers, calculators~\cite{cobbe2021training}, translation systems. 
We are the first to utilize this tool usage/function calling capability to solve the challenging DST task in TOD with LLMs, bridging the gap between general conversation and task-oriented dialogues.

\begin{figure*}[!ht]
\begin{center}
\includegraphics[width=1\linewidth]{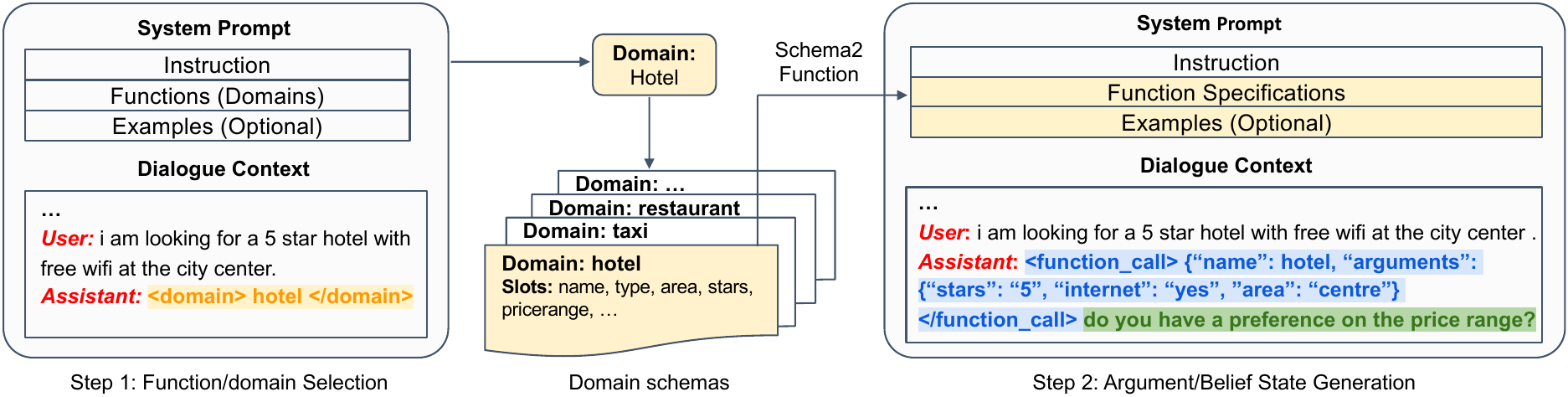}
\end{center}
   \caption{\textbf{Overview of our approach that addresses DST via function calling.} The whole prompt includes two sections: \textit{system prompt} and \textit{dialogue context}. The function calling process is executed in two stages. Initially, the model is prompted to determine \colorbox{yellow}{the function to be called (function name)}. Following this, the specifications of the predicted function/domain, along with optional demonstration examples, are incorporated into the \textit{system prompt}. This guides the model to generate \colorbox{myblue1}{function arguments} and subsequently the \colorbox{mygreen1}{response}. 
   }\label{fig:overview}
\end{figure*}

\section{Background}
\subsection{Chat-tuned LLMs}
Chat-tuned LLMs are models specifically fine-tuned to interact with users in a conversational manner. This category encompasses proprietary models such as ChatGPT and Claude, as well as open-source models such as Vicuna~\cite{vicuna2023}, LLaMA2-Chat~\cite{llama2}, and Baichuan~\cite{yang2023baichuan}. These chat models typically start as base models that are further fine-tuned with a dialogue format, enabling them to function effectively as conversational agents. As depicted in Figure~\ref{fig:overview}, the dialogue format typically features three distinct roles within two components: (1) the \textit{system} role in the \textit{system prompt} section, which defines the assistant's roles, responsibilities, and expected behaviors; and (2) the \textit{user} and \textit{assistant} roles in the \textit{dialogue context} section, encompassing their conversation. The model is typically tasked to produce the \textit{assistant}'s responses to the \textit{user}'s input. These chat models are primarily designed to generate helpful, detailed, and friendly responses to general user inquiries, rather than handling task-specific conversations as in TOD.

\subsection{DST Task Formulation}
In TOD, at each turn of conversation, the task of DST is to summarize the dialogue state $S_t$ given the dialogue context $C_t=\{A_1, U_1, \cdots, A_t, U_t\}$, where $U_t$ and $A_t$ represent the user utterance and assistant response at the $t$-th turn. For simplicity, we will omit the turn index $t$ in subsequent discussions. The dialogue state $S$ is a set of slot-value pairs:
\begin{equation}
    S = \{(s_{1,D_1}, v_{1,D_1}), \cdots, (s_{i,D_j}, v_{i,D_j})\},
\end{equation}
where $s_{i,D_j}$ is the $i$-th slot in the $D_j$ domain, and $v_{i,D_j}$ is its tracked value.
Each domain $D_j$ corresponds to a set of slots for a specific service, API call, or database query, such as restaurant reservations. In the case of the \textit{restaurant} domain, the slots might include ``restaurant-food'', ``restaurant-area'', ``restaurant-pricerange'', etc. We use $S_{D_j}$ to denote the tracked slots for domain $D_j$.

\begin{figure}[!t]
\begin{center}
\includegraphics[width=1\linewidth]{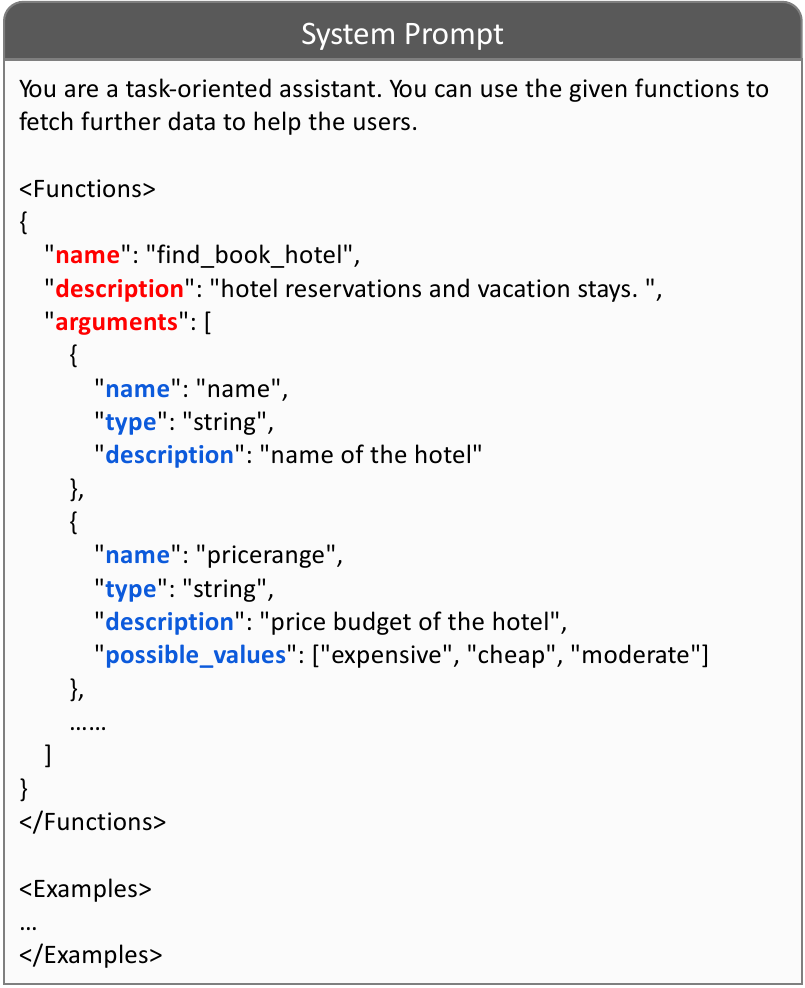}
\end{center}
   \caption{Illustration of the \textit{system prompt} consisting of three components: (1) the overall instruction, (2) function specifications, and (3) optional example conversations. Due to space constraints, only a part of slots/arguments are displayed. The specific example conversations are omitted for brevity.
   }\label{fig:system}
\vspace{-3mm}
\end{figure}

\section{Approach}
\begin{figure}[!t]
\begin{center}
\includegraphics[width=1\linewidth]{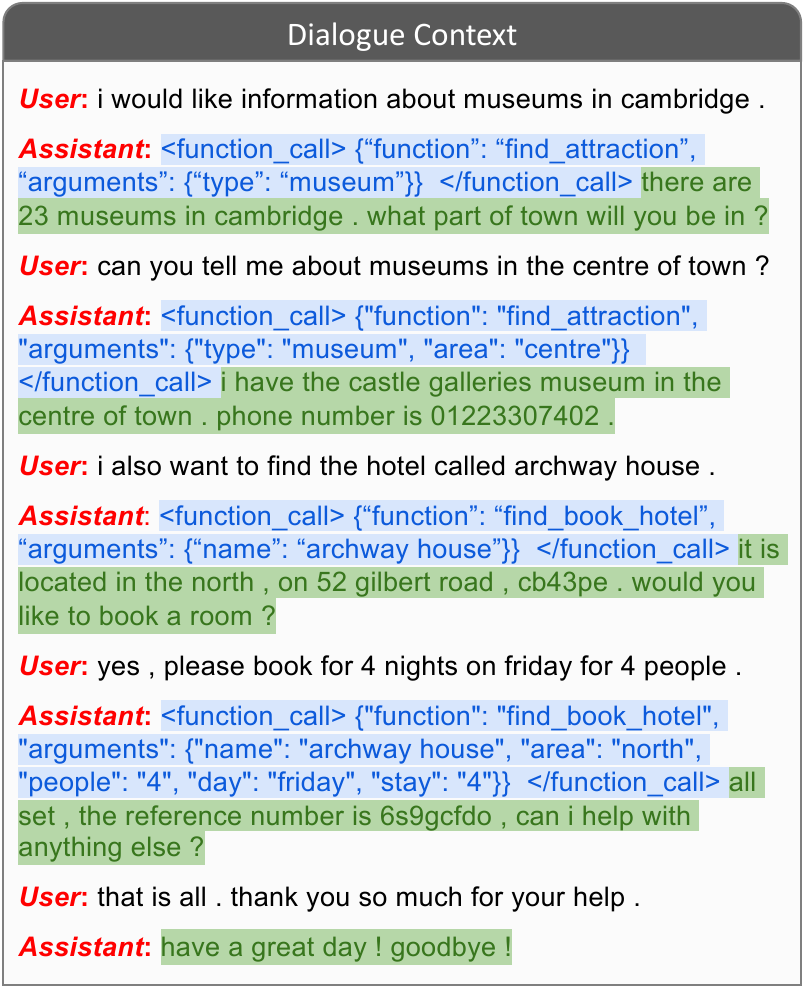}
\end{center}
   \caption{An example of the \textit{dialogue context} including attraction and hotel domains. 
   The assistant output consists of the \colorbox{myblue1}{function calls} and \colorbox{mygreen1}{response}. 
   }\label{fig:conversation}
\vspace{-3.5mm}
\end{figure}

\vspace{-2mm}
Our method redefines DST as function calling, treating each domain as a distinct function, and the slot values within the domain as its arguments. As shown in Figure~\ref{fig:overview}, this paradigm is represented in chat-tuned models by embedding function specifications within system prompts, as shown in Figure~\ref{fig:system}. 
The model is tasked with generating function calls followed by a response, as shown in Figure~\ref{fig:conversation}. We provide more details of our approach below.

\vspace{-0.7mm}
\paragraph{DST as Function Calling}
In our formalization, DST is conceptualized as function calling. Each domain $D_j$ is modeled as a unique function $F_j$, with the associated slot values serving as arguments. Consequently, at each turn of the conversation, the DST task transforms into identifying the correct function $F_j$ and its arguments $S_{D_j}$:
\begin{equation}
\begin{aligned}
    \texttt{<fn\_call>} \: F_j(s_{1,D_1}=v_{1,D_1}, s_{2,D_1}=v_{2,D_1}, \\
    \cdots, s_{i,D_j}=v_{i,D_j}) \: \texttt{</fn\_call>},
\end{aligned}
\end{equation}
where ``\texttt{<fn\_call>}'' and ``\texttt{</fn\_call>}'' are special tokens. In practice, we use ``\texttt{<function\_call>}'' and ``\texttt{</function\_call>}'' to represent them, and generate the function calls in JSON format. Some examples of function calls generated within a conversation are shown in Figure~\ref{fig:conversation}.

\vspace{-0.7mm}
\paragraph{Dialogue Prompt Format}
As shown in Figure~\ref{fig:conversation}, we incorporate this function calling as an integral part of the conversation. At each turn of the conversation, given the current conversation context, the chat model is tasked with first generating the function call, followed by the response. 
To achieve that, we convert the domain schema into function specifications, using JSON, and include them in the \textit{system prompt} within the dialogue prompt, as shown in Figure~\ref{fig:system}.
By doing so, the model can be aware of the supported functions and the correct generation format to call the function.

\vspace{-0.7mm}
\paragraph{Function Call Decomposition}
As outlined, the model is tasked with predicting not only the function to call (\textit{i.e.}, function name) but also generating arguments for the predicted function. Providing detailed specifications for all the supported functions would make the prompt very lengthy. To streamline this process and minimize the prompt length, we split the whole generation process into two consecutive steps: \textbf{Function Selection} and \textbf{Argument Generation}. As shown in Figure~\ref{fig:overview}, during each turn of the conversation, the model first selects a function $F_j$ from all supported functions. In this step, we provide concise function descriptions rather than detailed specifications in the \textit{system prompt} and instruct the model to generate only the selected domain/function, surrounded by the special tokens ``\texttt{<domain>}'' and ``\texttt{</domain>}''. Subsequently, we include the detailed specification of only the chosen function $F_j$ in the \textit{system prompt}, prompting the model to generate the corresponding arguments $F_j$.
This approach not only simplifies the task but also significantly reduces the prompt length, enhancing both efficacy and efficiency.

\paragraph{In-context Prompting}
Since the current open-source models are not specifically fine-tuned to generate function calls, there is no guarantee that the model could always generate the correct formats. To address that, we also include in-context example conversations as shown in Figure~\ref{fig:conversation}, along with the specification of the predicted function in the \textit{system prompt}. We manually selected a few demonstration examples for each supported domain.

\paragraph{Prompt-based Fine-tuning}
To illustrate equipping an open source model with function calling capabilities, eliminating the need for demonstration examples, we fine-tune a \textsc{LLaMA2-13B-Chat} model using a collection of heterogeneous task-oriented dialogue datasets, including WOZ~\cite{woz}, CamRest676~\cite{camres6761,camres6762}, MSR-E2E~\cite{mse2e}, TaskMaster~\cite{taskmaster} and Schema-Guided Dialogues (SGD)~\cite{sgd}. Note that we deliberately exclude the entire target dataset.
From these datasets, we choose 36 distinct domains with high-quality annotations. Instead of using all the data in those datasets, we randomly sampled 200 dialogues from each domain across the datasets, totaling 7,200 dialogues for training. This small sample size proved sufficient. 

We convert these dialogues into our designed dialogue prompt. Specifically, we incorporate the specifications of all functions invoked in each conversation's \textit{system prompt}. Our loss calculation focused solely on the function calling aspect of the assistant's generation. We refrained from fine-tuning the response generation component, in consideration of the LLMs' existing competence in producing coherent responses and the scarcity of function-calling examples in our dataset. 
Our fine-tuned model is dubbed \textsc{FnCTOD-LLaMA2-13B}.

\begin{table*}[!ht]
\centering
\resizebox{1.0\linewidth}{!}{
\begin{tabular}{lcccccccccccc}
\toprule
\multirow{2}{*}{\textbf{Model}} & \multicolumn{2}{c}{Attraction} & \multicolumn{2}{c}{Hotel} & \multicolumn{2}{c}{Restaurant} & \multicolumn{2}{c}{Taxi} & \multicolumn{2}{c}{Train} & \multicolumn{2}{c}{\textbf{JGA}} \\
		\cmidrule(lr){2-3}
		\cmidrule(lr){4-5}
		\cmidrule(lr){6-7}
		\cmidrule(lr){8-9}
  		\cmidrule(lr){10-11}
		\cmidrule(lr){12-13}
& \textbf{JGA} & \textbf{Slot-F1} & \textbf{JGA} & \textbf{Slot-F1} & \textbf{JGA} & \textbf{Slot-F1} & \textbf{JGA} & \textbf{Slot-F1} & \textbf{JGA} & \textbf{Slot-F1} & \textbf{Average} & \textbf{Overall} \\
\midrule
\rowcolor{Gray}
\multicolumn{13}{c}{\textit{\textbf{\large Cross-domain Transfer approaches}}} \\
TRADE~\cite{trade} & 20.06 & -- & 14.20 & -- & 12.59 & -- & 59.21 & -- & 22.39 & -- & 25.69 & -- \\ 
MA-DST~\cite{ma-dst} & 22.46 & -- & 16.28 & -- & 13.56 & -- & 59.27 & -- & 22.76 & -- & 26.87 & -- \\ 
TransferQA~\cite{lin-etal-2021-zero} & 31.25 & -- & 22.72 & -- & 26.28 & -- & 61.87 & -- & 36.72 & -- & 35.77 & --  \\
T5DST~\cite{lin-etal-2021-leveraging} & 33.09 & -- & 21.21 & -- & 21.65 & -- & 64.62 & -- & 35.43 & -- & 35.20 & -- \\ 
D3ST~\cite{D3ST} & 56.40 & -- & 21.80 & -- & 38.20 & -- & \textbf{78.40} & -- & 38.70 & -- & 46.70 & -- \\ 
\midrule
\rowcolor{Gray}
\multicolumn{13}{c}{\textit{\textbf{\large Previous Prompting approaches}}} \\
*IC-DST (Codex) & 60.00 & -- & 46.70 & -- & 57.30 & -- & 71.40 & -- & 49.40 & -- & 56.96 & --\\
\citet{heck2023chatgpt} (GPT-3.5) & 52.70 & -- & 42.00 & -- & 55.80 & -- & 70.90 & -- & 60.80 & -- & 56.44 & 31.50\\
InstructTODS (GPT-3.5) & 30.23 & 65.38 & 26.77 & 76.28 & 48.28 & 82.90 & 56.22 & 75.33 & 53.75 & 83.64 & 42.02 & --\\
InstructTODS (GPT-4) & 39.53 & 78.99 & 31.23 & 84.07 & 55.86 & 88.23 & 63.24 & 82.71 & 59.83 & 89.72 & 48.16 & -- \\
\midrule
\rowcolor{Gray}
\multicolumn{13}{c}{\textit{\textbf{\large
 Our approach \textsc{FnCTOD}}}} \\
\rowcolor{mygreen1}
ChatGPT (GPT-3.5) & \textbf{67.15} & \textbf{87.20} & 37.56 & 82.86 & 60.12 & 90.21 & 74.43 & 86.90 & 67.29 & \textbf{92.48} & 61.31 & 38.56\\
\rowcolor{mygreen1}
ChatGPT (GPT-4) & 58.77 & 81.84 & 45.15 & 85.07 & 63.18 & 91.06 & 76.39 & \textbf{87.73} & \textbf{69.48} & 90.16 & \textbf{62.59} & \textbf{38.71}\\
\rowcolor{lightcyan}
\textsc{FnCTOD-LLaMA2-13B} & 62.24 & 84.99 & \textbf{46.83} & \textbf{85.39} & 60.27 & 88.69 & 67.48 & 80.39 & 60.90 & 89.88 & 59.54 & 37.67\\

\rowcolor{lightyellow}
\textsc{Zephyr-7B-Beta} & 56.50 & 81.97 & 38.43 & 79.52 & \textbf{63.18} & \textbf{91.19} & 74.10 & 86.56 & 56.20 & 90.00 & 57.68 & 32.11\\
\rowcolor{lightyellow}
\textsc{Vicuna-7B-v1.5} & 50.66 & 74.93 & 35.00 & 73.66 & 52.76 & 85.25 & 67.02 & 80.36 & 59.66 & 89.05 & 53.02 & 29.45 \\
\rowcolor{lightyellow}
\textsc{Vicuna-13B-v1.5} & 54.25 & 80.99 & 38.43 & 79.96 & 56.44 & 87.26 & 69.11 & 83.37 & 58.82 & 89.26 & 55.41 & 31.84 \\
\rowcolor{lightyellow}
\textsc{Baichuan2-13B-Chat} & 53.67 & 79.57 & 40.15 & 81.36 & 59.02 & 87.82 & 69.31 & 81.95 & 60.67 & 89.45 & 56.56 & 33.21 \\
\rowcolor{lightyellow}
\textsc{LLaMA2-7B-Chat} & 42.64 & 70.18 & 30.47 & 69.37 & 37.60 & 78.63 & 63.20 & 73.80 & 44.17 & 82.18 & 43.44 & 16.78 \\
\rowcolor{lightyellow}
\textsc{LLaMA2-13B-Chat} & 49.76 & 76.80 & 29.50 & 67.60 & 48.87 & 81.33 & 64.66 & 68.97 & 53.59 & 85.09 & 49.28 & 25.68 \\
\rowcolor{lightyellow}
\textsc{LLaMA2-70B-Chat} & 50.66 & 78.26 & 34.03 & 76.61 & 54.48 & 86.18 & 66.10 & 72.60 & 56.53 & 87.39 & 52.36 & 28.38 \\
\bottomrule
\end{tabular}
}
\caption{\textbf{Performance comparison on zero-shot DST benchmark}. We compare our approach with cross-domain approaches and prompting approaches relying on ChatGPT (GPT-3.5/4) and Codex. 
Using our approach, we evaluate \colorbox{mygreen1}{ChatGPT}, and \colorbox{lightcyan}{our fine-tuned model} via zero-shot prompting, and \colorbox{lightyellow}{open-source models} via few-shot (5-shot) prompting.
In addition to per-domain JGA and slot F1 scores, we report the macro-averaged JGA of these five domains (\textbf{Average JGA}), and also the multi-domain JGA (\textbf{Overall JGA}).
The baseline results are directly taken from their respective works. The best performances in each column are in \textbf{bold}.}
\label{tab:dst_result}
\vspace{-4.75mm}
\end{table*}

\vspace{-2mm}
\section{Experiments}\label{sect:experiments}

\vspace{-1mm}
\subsection{Experimental Setup}
\paragraph{Dataset and Metrics}
    We evaluate on the widely-used task-oriented multi-domain dataset \textbf{MultiWOZ 2.1}~\cite{budzianowski-etal-2018-multiwoz,eric-etal-2020-multiwoz}. We used the 1,000 dialogues in the test split and measured joint goal accuracy (JGA), which measures the percentage of turns for which all slot values are correctly predicted.
    This test set spans 5 domains, with each conversation potentially covering multiple domains.

\vspace{-1.75mm}
\paragraph{Baselines}
We compare our approach with two distinct approaches: (1) \textit{Cross-domain transfer approaches}, which involve training on MultiWOZ with one domain excluded and then evaluating on the held-out domain.
This category includes TRADE~\cite{trade}, MA-DST~\cite{ma-dst}, TransferQA~\cite{lin-etal-2021-zero}, T5DST~\cite{lin-etal-2021-leveraging}, and D3ST~\cite{D3ST}. 
(2) \textit{Previous prompting approaches} that have only shown efficacy with advanced proprietary models, include IC-DST~\cite{icdst} using Codex, \cite{heck2023chatgpt} and InstructTODS~\cite{instructtods} using ChatGPT (GPT-3.5/4). 

{\let\thefootnote\relax\footnotetext{\hspace{-1.5mm}*\hspace{0.2mm}IC-DST requires in-domain data to train the retriever for example selection, making it not strictly zero-shot DST.}}

\vspace{-1.5mm}
\paragraph{Evaluated Models}
We evaluate our method on proprietary ChatGPT and various open-source models. For ChatGPT, we evaluated the versions of GPT-3.5-Turbo (\texttt{gpt-3.5-turbo-1106}) and GPT-4 (\texttt{gpt-4-1106-preview}), both of which are already equipped with function calling capabilities.
Regarding open-source models, we assessed several widely recognized chat-tuned models of varying sizes, including the 7B parameter model \textsc{Zephyr-7B-Beta}~\citep{zephyr}, the 7B and 13B versions of \textsc{Vicuna-v1.5}~\citep{vicuna2023}, the 7B, 13B, and 70B versions of \textsc{LLaMA2-Chat}~\citep{llama2}, as well as the 13B parameter model \textsc{Baichuan2-13B-Chat}~\cite{baichuan2}.

Additionally, we evaluate our fine-tuned model \textsc{FnCTOD-LLaMA2-13B}. It's worth noting that unlike these domain transfer baselines, our model is trained exclusively on 7,200 dialogues from datasets other than MultiWOZ, making the setup more realistic and challenging.

\vspace{-3mm}
\paragraph{Inference Details}
For both ChatGPT and our fine-tuned \textsc{FnCTOD-LLaMA2-13B}, which have been equipped with function-calling capabilities, we perform zero-shot prompting, excluding in-context examples in the system prompt. For the other open-source models, we perform few-shot prompting using five examples (5-shot) by default. It's worth noting that the shot in zero/few-shot prompting refers to the number of in-context examples used when prompting the models, whereas the shot in zero-shot DST refers to the number of in-domain examples seen in the training data. 
%

\vspace{-1.5mm}
\subsection{Zero-shot DST Evaluation}
\vspace{-1mm}
Table~\ref{tab:dst_result} presents the zero-shot DST performance comparison, with observations summarized below.


\paragraph{Our approach empowers moderately-sized open-source models to surpass previous SOTA results achieved with advanced ChatGPT.}
Previous prompting approaches showed promising results exclusively with advanced proprietary models but underperformed with less advanced models~\cite{hudevcek2023llms}. 
Our approach is the first to enable moderately sized open-source models to achieve comparable or superior performance compared to previous SOTA results obtained with advanced ChatGPT and Codex. Specifically, the 7B parameter \textsc{Zephyr-7B-Beta} and 13B parameter \textsc{Baichuan2-13B-Chat} models outperform the previous SOTA. This significant advancement marks a milestone in the practical application of LLMs for DST and TOD.

\paragraph{Our approach significantly improves ChatGPT's performance over previous prompting approaches.}
The efficacy of our approach is verified by improvements of 4.8\% (Average JGA) for GPT-3.5, and 14\% for GPT-4, compared to previous reported results with each of these models. Our result with GPT-4 beats the previous SOTA prompting approach using Codex by 5.6\% Avergage JGA.

\paragraph{Our fine-tuned 13B parameter model matches the performance of ChatGPT.} 
It is evident that our fine-tuned \textsc{FnCTOD-LLaMA2-13B} significantly improves over its base model \textsc{LLaMA2-13B-Chat} and achieves a performance comparable to ChatGPT. This demonstrates that we can easily equip moderately sized open-source LLMs with function-calling capabilities and zero-shot DST performance comparable to ChatGPT, marking an exciting advance in bridging the gap between open-source and proprietary models.

\subsection{Zero-shot End-to-End TOD Evaluation}

In practical settings, a TOD system queries a knowledge base or API using the tracked dialogue states to ground responses.
We perform an end-to-end evaluation of both DST and response generation, which is a more realistic and challenging setting.
Our \textsc{FnCTOD} approach enables the generation of both dialogue states, \textit{i.e.}, function calls, \emph{and} responses in the assistant's output. This contrasts with the prompting methods that typically treat DST as a standalone task. Consistent with the previous work on end-to-end zero-shot TOD evaluation~\cite{hudevcek2023llms}, we evaluated using the MultiWOZ 2.2 dataset~\cite{multiwoz22} with delexicalized responses. Our evaluation metrics include JGA for DST and \textit{Success} rate for the generated response. Success measures the percentage of dialogues in which the user's goals were fully met \cite{budzianowski-etal-2018-multiwoz}. The results are presented in Table~\ref{tab:nlg}.

\begin{table}[!ht]
\centering
\resizebox{1\linewidth}{!}{
\begin{tabular}{lcc}
\toprule
\textbf{Model} & \textbf{JGA} & \textbf{Success} \\ \midrule

ChatGPT~\cite{hudevcek2023llms} & 27.0 & 44.0 \\ \midrule

\rowcolor{lightcyan}
\textsc{FnCTOD-LLaMA2-13B} & \textbf{37.9} & 44.4 \\

\rowcolor{lightyellow}
\textsc{Zephyr-7B-Beta} & 32.3 & \textbf{57.5} \\
\rowcolor{lightyellow}
\textsc{Vicuna-7B-v1.5} & 29.4 & 37.7 \\
\rowcolor{lightyellow}
\textsc{Vicuna-13B-v1.5} & 33.8 & 23.1 \\
\rowcolor{lightyellow}
\textsc{Baichuan2-13B-Chat} & 33.0 & 45.7 \\
\rowcolor{lightyellow}
\textsc{LLaMA2-7B-Chat} & 16.7 & 24.9 \\
\rowcolor{lightyellow}
\textsc{LLaMA2-13B-Chat} & 25.8 & 27.7 \\
\bottomrule
\end{tabular}
}
\caption{End-to-end evaluation results on MultiWOZ 2.2, including the evaluation on DST with (overall) \textbf{JGA} and also response generation with \textbf{Success rate}.}
\label{tab:nlg}
\vspace{-3mm}
\end{table}

Compared to previous prompting approaches, by enabling both zero-shot DST and response generation~\cite{hudevcek2023llms}, the superiority of the FnCTOD approach becomes more evident. Specifically, all open-source models evaluated using our approach outperform ChatGPT's results achieved by~\cite{hudevcek2023llms}, except for \textsc{LLaMA2-7B-Chat}.
In addition, the results show that the fine-tuned model \textsc{FnCTOD-LLaMA2-13B} retains its ability to generalize and generate informative responses in a zero-shot TOD setting.

\subsection{Ablation Studies}\label{sect:analysis}



\begin{table*}[!ht]
\centering
\resizebox{1.0\linewidth}{!}{
\begin{tabular}{lccccccccccccc}
\toprule
\multirow{2}{*}{\textbf{Method}} & \multicolumn{6}{c}{\textbf{JGA} ($\uparrow$)} & \multicolumn{3}{c}{\textbf{Tokens (m) ($\downarrow$)}} & \multicolumn{3}{c}{\textbf{Time (h) ($\downarrow$)}} 
& \multicolumn{1}{c}{\textbf{API} ($\downarrow$)} \\
		\cmidrule(lr){2-7}
		\cmidrule(lr){8-10}
		\cmidrule(lr){11-13}
  \cmidrule(lr){14-14}
& Attr. & Hotel & Rest. & Taxi & Train & Overall & FS & AG & Total & FS & AG & Total & \textbf{Cost (\$)}\\
\midrule
\rowcolor{Gray}
\multicolumn{14}{c}{\textbf{ChatGPT (GPT-3.5)}} \\
Non-decomp. & 59.64 & 32.24 & 61.39 & 74.87 & 49.91 &30.16 &-&-&-&-&-&-&13.32\\ 
Decomposition & \textbf{67.15} & \textbf{37.56} & 60.12 & 74.43 & \textbf{67.29} &\textbf{38.56}&-&-&-&-&-&-&\textbf{8.97}\\ 
\rowcolor{Gray}
\multicolumn{14}{c}{\textit{\textbf{\textsc{FnCTOD-LLaMA2-13B}}}} \\
Non-decomp. & 34.77 & 32.02 & 56.63 & 65.40 & 36.21 &21.04 &-&-&23.57&-&-&12.53&-\\ 
Decomposition & \textbf{62.24} & \textbf{46.83} & \textbf{60.27} & \textbf{67.48} & \textbf{60.90} & \textbf{37.67} &5.24&7.77&\textbf{13.01}&0.80&8.64&\textbf{9.44}&-\\ 
\bottomrule
\end{tabular}
}
\caption{Ablation studies on the function call generation decomposition, where decomp. denotes decomposition. The comparison on efficiency includes the API cost of \textsc{GPT-3.5} and local inference time and consumed tokens for \textsc{FnCTOD-LLaMA2-13B} on a single Nvidia A6000 GPU, where FS stands for \underline{F}unction \underline{S}election (the first step) and AG stands for \underline{A}rgument \underline{G}eneration (the second step). Best results are bolded.}
\label{tab:zs_ablation}
\vspace{-3mm}
\end{table*}


\paragraph{Impact of function call decomposition}



Our two-step decomposition approach can yield benefits in terms of both efficacy and efficiency. To demonstrate this, we compared the accuracy and cost with and without the two-step decomposition. The ``without decomposition'' condition includes the full specification of all domains/functions in the prompt, and requires the LLM to generate the full function call, including the function name and arguments in one step.
This comparison is conducted on ChatGPT and our fine-tuned \textsc{FnCTOD-LLaMA2-13B}, which supports zero-shot prompting.
Since ChatGPT is API-accessible, we compare its API calling cost which is calculated based on the number of tokens in the prompt.
The results in Table~\ref{tab:zs_ablation} demonstrate both the effectiveness and efficiency of our decomposition approach.

\begin{table}[h!]
\centering
\resizebox{1.0\linewidth}{!}{
\begin{tabular}{lccc}
\toprule
\textbf{Model} & \textbf{Oracle} & \textbf{FS Acc.} & \textbf{JGA} \\
\midrule

\rowcolor{lightgrey}
ChatGPT (GPT-4) & $\times$ & 88.62 & 38.71\\
\rowcolor{lightgrey}
ChatGPT (GPT-4) & $\checkmark$ & - & 44.44 \\ 

ChatGPT (GPT-3.5) & $\times$ & 95.54 & 38.56 \\
ChatGPT (GPT-3.5) & $\checkmark$ & - & 38.32 \\ \hline

\rowcolor{lightgrey}
\textsc{FnCTOD-LLaMA2-13B} & $\times$ & 91.68 & 37.67 \\
\rowcolor{lightgrey}
\textsc{FnCTOD-LLaMA2-13B} & $\checkmark$ & - & 37.93\\

\textsc{Zephyr-7B-Beta} & $\times$ & 92.77 & 32.11\\
\textsc{Zephyr-7B-Beta} & $\checkmark$ & - & 34.40\\ 

\rowcolor{lightgrey}
\textsc{Vicuna-7B-v1.5} & $\times$ & 94.75 & 29.45 \\
\rowcolor{lightgrey}
\textsc{Vicuna-7B-v1.5} & $\checkmark$ & - & 30.06 \\

\textsc{Vicuna-13B-v1.5} & $\times$ & 91.82 & 31.84 \\
\textsc{Vicuna-13B-v1.5} & $\checkmark$ & - & 34.20 \\ 

\rowcolor{lightgrey}
\textsc{Baichuan2-13B-Chat} & $\times$ & 92.50 & 33.21 \\
\rowcolor{lightgrey}
\textsc{Baichuan2-13B-Chat} & $\checkmark$ & - & 34.93 \\

\textsc{LLaMA2-7B-Chat} & $\times$ & 91.90 & 16.78 \\
\textsc{LLaMA2-7B-Chat} & $\checkmark$ & - & 18.75 \\

\rowcolor{lightgrey}
\textsc{LLaMA2-13B-Chat} & $\times$ & 89.34 & 25.68 \\
\rowcolor{lightgrey}
\textsc{LLaMA2-13B-Chat} & $\checkmark$ & - & 26.56 \\


\bottomrule
\end{tabular}
}
\caption{Model performances with oracle and predicted functions/domains. FS Acc. indicates the function selection accuracy, and the JGA here indicates overall JGA.}
\label{tab:dp}
\vspace{-5mm}
\end{table}

In addition, to investigate the impact of cascading, two-stage errors due to incorrect function (name) prediction, we conduct experiments to compare the performance with oracle and predicted functions. The results are shown in Table~\ref{tab:dp}. As can be seen, while errors do impact the overall performance, the significance depends on the prediction accuracy. Many LLMs, especially ChatGPT (GPT-3.5), demonstrate very high prediction accuracy considering potential label noise. Performance with oracle and predicted functions are largely comparable. Improving the function prediction accuracy could further boost the performance of our approach. 

\begin{figure*}[!ht]
  \centering
  \subfigure{
  \begin{minipage}[b]{0.33\textwidth}
    \includegraphics[width=\linewidth]{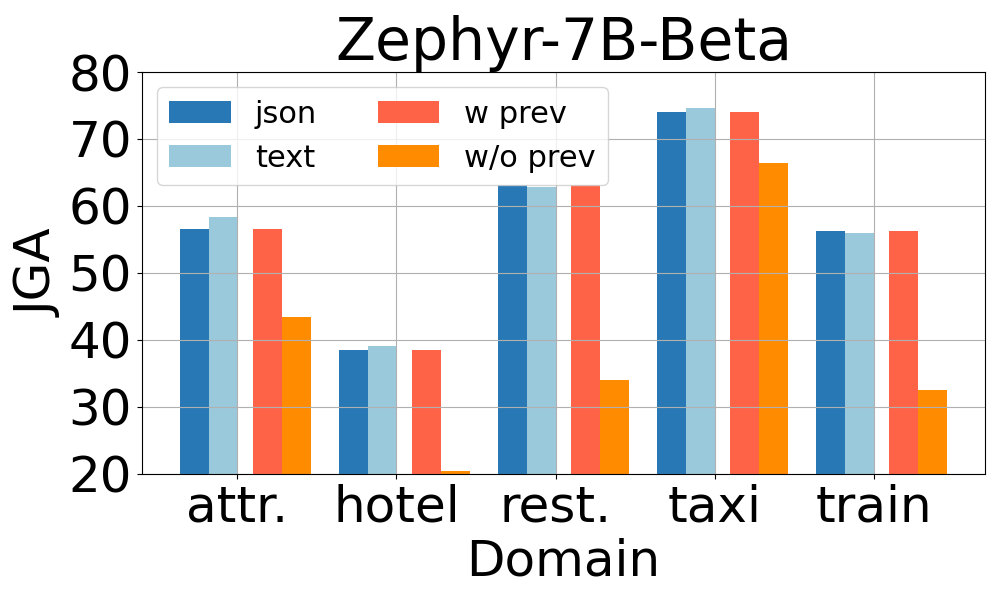}
    \end{minipage}
    \begin{minipage}[b]{0.33\textwidth}
        \includegraphics[width=\linewidth]{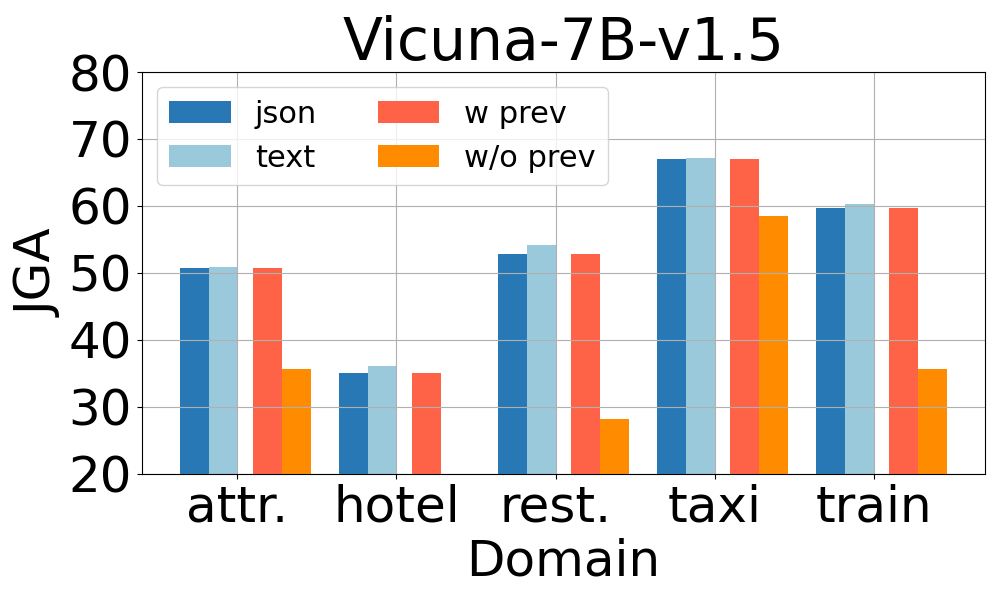}
    \end{minipage}
    \begin{minipage}[b]{0.33\textwidth}
        \includegraphics[width=\linewidth]{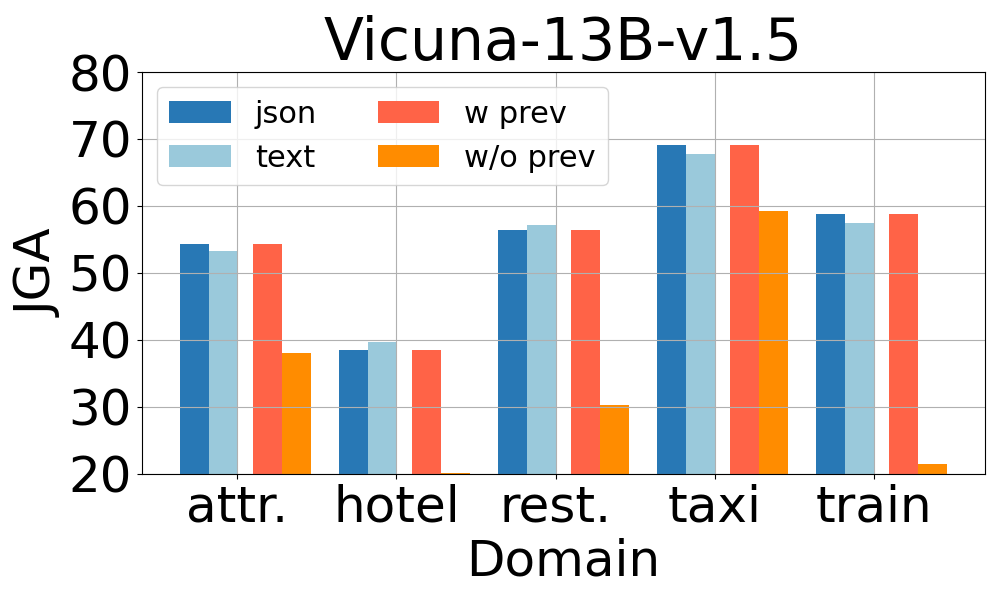}
    \end{minipage}
 }
 \vspace{-3mm}
  \subfigure{
  \begin{minipage}[b]{0.33\textwidth}
    \includegraphics[width=\linewidth]{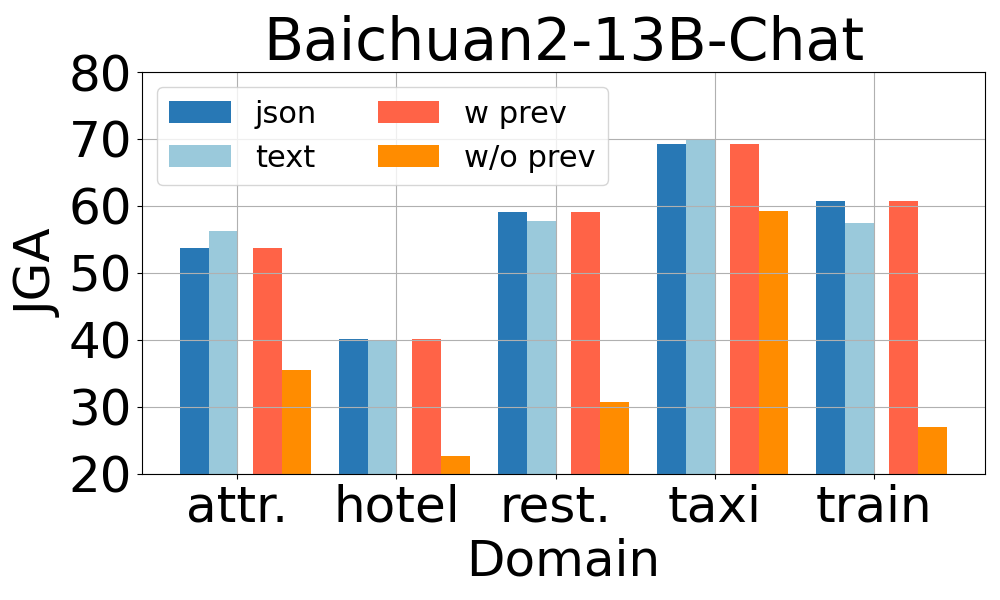}
    \end{minipage}
    \begin{minipage}[b]{0.33\textwidth}
        \includegraphics[width=\linewidth]{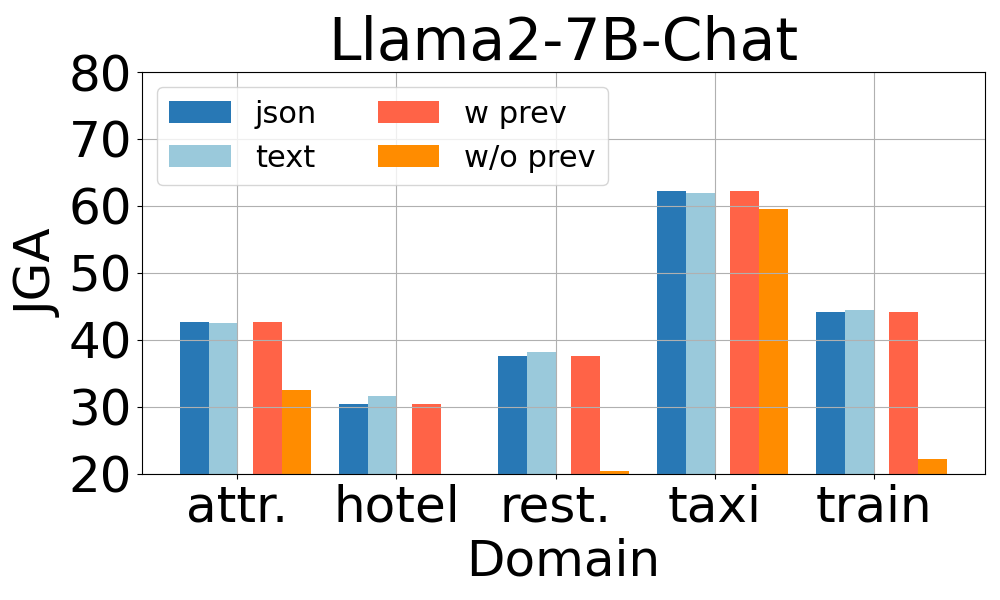}
    \end{minipage}
    \begin{minipage}[b]{0.33\textwidth}
        \includegraphics[width=\linewidth]{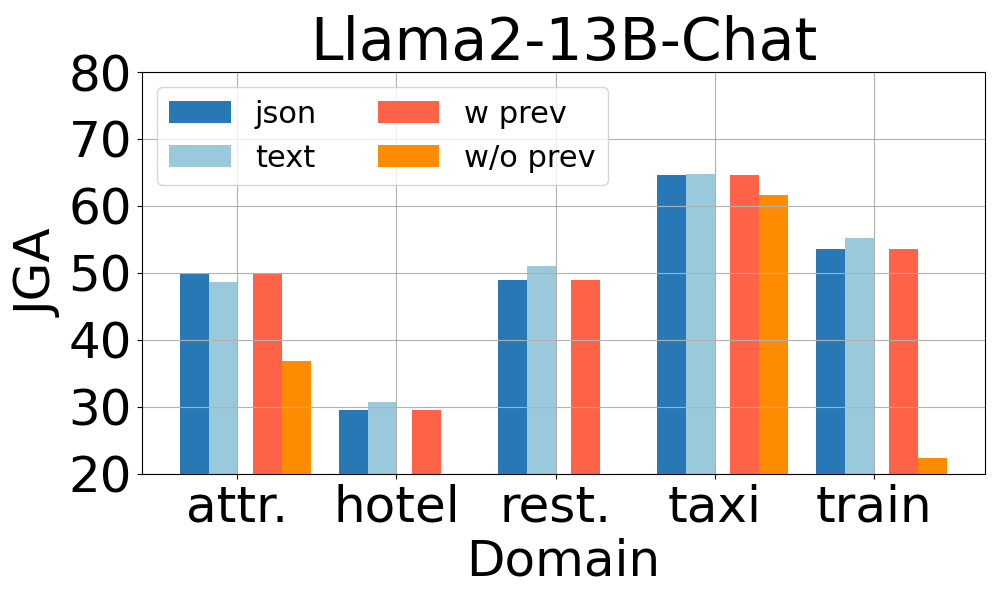}
    \end{minipage}
 }
  \caption{Ablation studies on different function specification types (json/text) and the unified dialogue format including or not including function calls in previous conversation context (w/ and w/o prev).
  }\label{fig:fs_ablation}
  \vspace{-3mm}
\end{figure*}

\paragraph{Impact of function specifications}
In addition to directly including function specifications in JSON within the prompt, we experimented with translating the data into more human-readable natural language descriptions (refer to the comparison in Figure~\ref{fig:type} in the Appendix). Figure~\ref{fig:fs_ablation} presents a comparison between using the JSON format directly (json) and converting it into natural language descriptions (text). The results indicate that the models perform similarly with both methods of function specification, indicating the high degree of flexibility of our approach in function specification customization. 

\paragraph{Impact of the unified dialogue prompt}
In our approach, we seamlessly integrated function calls into the assistant's output in the conversation context, which could also serve as demonstration for the current turn's generation. To investigate its effect, we show the performance with and without function calls (w/ and w/o prev) in Figure~\ref{fig:fs_ablation}. The results emphasize the effectiveness of embedding function calls within the conversation context.

\begin{figure*}[!ht]
  \centering
    \vspace{3mm}
  \subfigure{
  \includegraphics[width=0.73\linewidth]{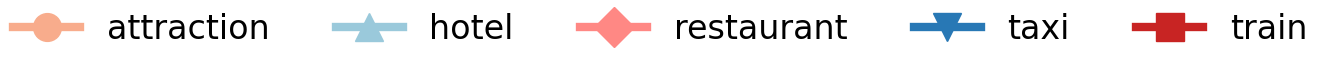}
 }
  \vspace{-2mm}
  \subfigure{
  \begin{minipage}[b]{0.32\textwidth}
    \includegraphics[width=\linewidth]{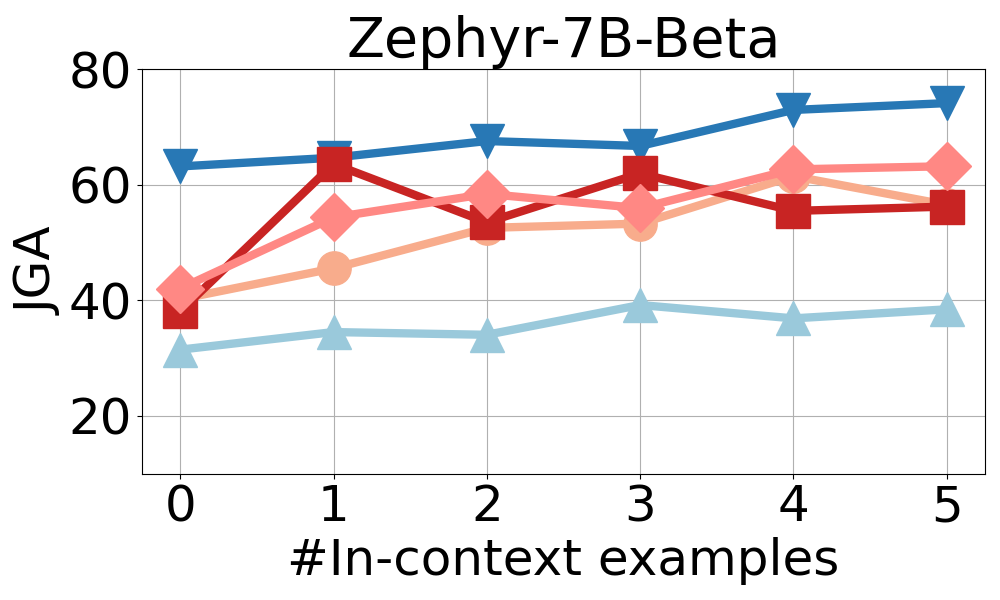}
    \end{minipage}
    \begin{minipage}[b]{0.32\textwidth}
        \includegraphics[width=\linewidth]{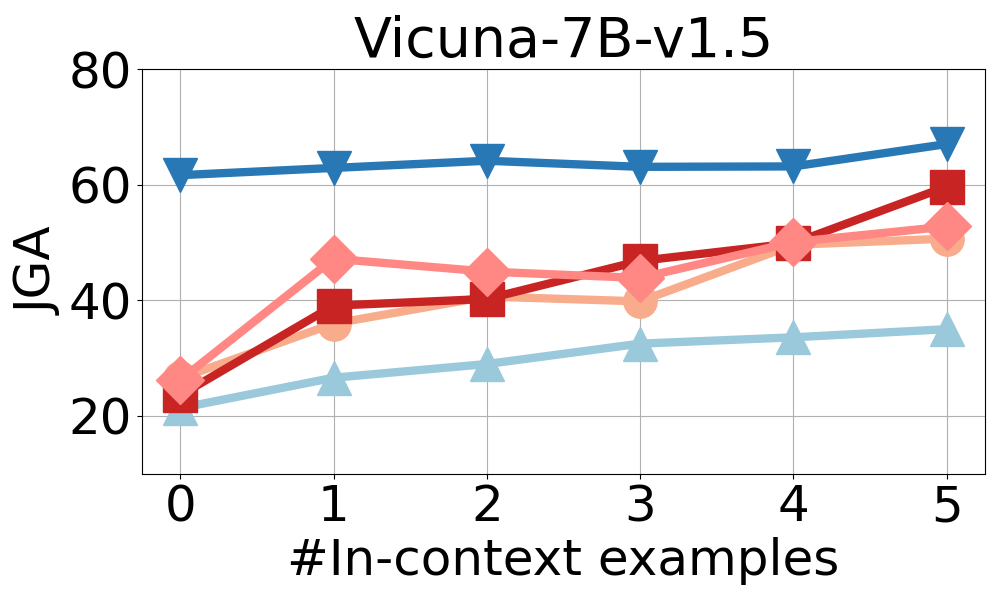}
    \end{minipage}
    \begin{minipage}[b]{0.32\textwidth}
        \includegraphics[width=\linewidth]{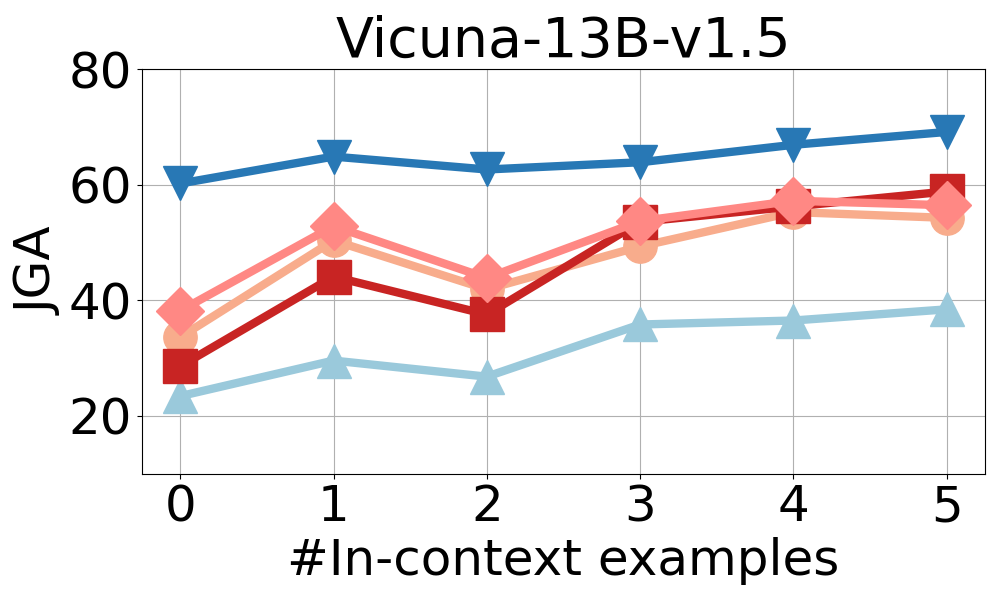}
    \end{minipage}
 }
 \vspace{-2mm}
  \subfigure{
  \begin{minipage}[b]{0.32\textwidth}
    \includegraphics[width=\linewidth]{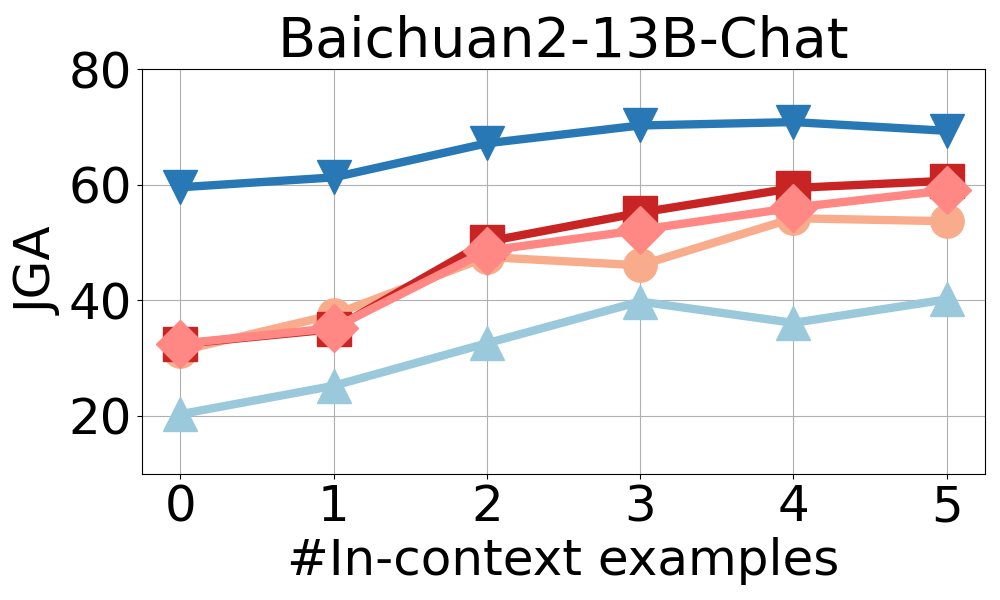}
    \end{minipage}
    \begin{minipage}[b]{0.32\textwidth}
        \includegraphics[width=\linewidth]{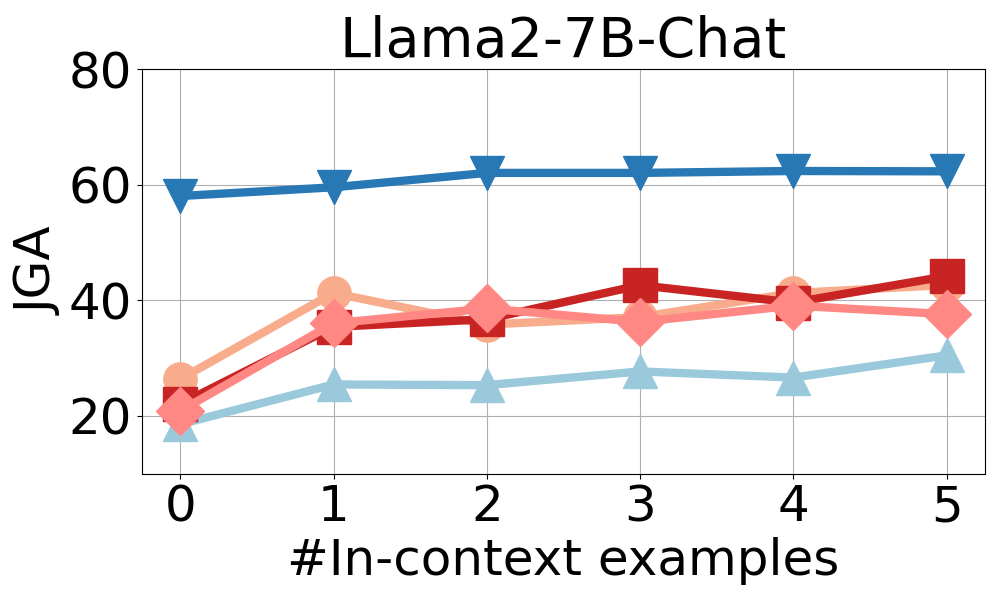}
    \end{minipage}
    \begin{minipage}[b]{0.32\textwidth}
        \includegraphics[width=\linewidth]{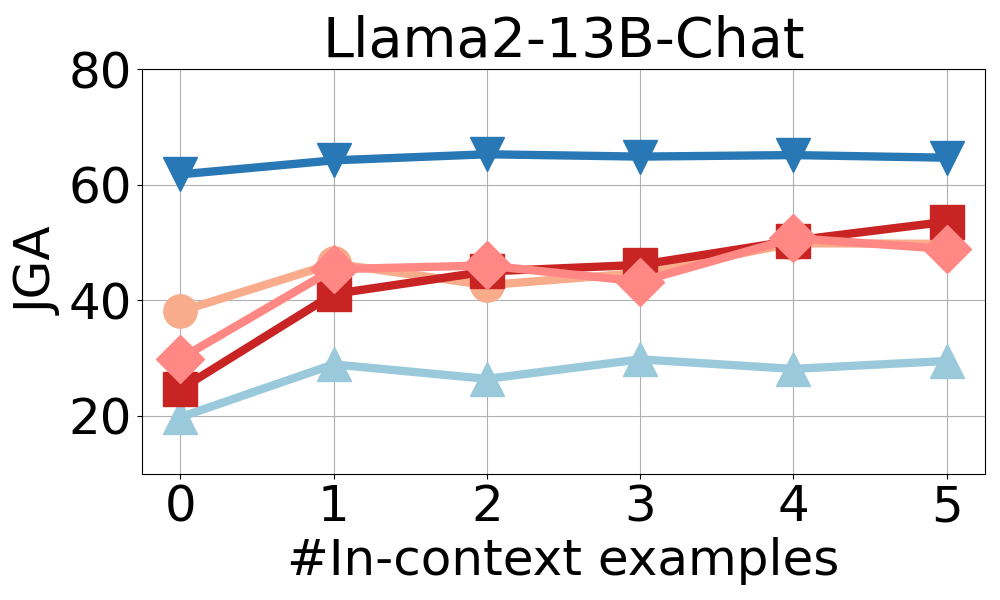}
    \end{minipage}
}
  \caption{Performance of open-source models with different numbers of in-context examples.
  }\label{fig:nshot}
  \vspace{-3mm}
\end{figure*}

\paragraph{Impact of varying numbers of in-context examples}
We assessed the performance of various open-source models, which were not originally trained for function call generation, with different numbers of in-context examples, ranging from 0 to 5. We note that using more than five examples might surpass the context-window capacity (such as 4096 tokens) for some models. The findings are illustrated in Figure~\ref{fig:nshot}.
The results indicate that the models perform significantly better when in-context examples are utilized compared to zero-shot prompting. There is a consistent performance improvement as the number of examples increases, across most domains and models. 

\paragraph{Impact of fine-tuning data sizes}
Our results indicate that with as few as 200 samples per domain, totaling 7,200 dialogues across 36 domains, we were able to fine-tune a \textsc{LLaMA2-13B-Chat} model to match the zero-shot DST performance of ChatGPT. We explored the model's performance with varying numbers of samples, ranging from 100 to 400 per domain. The results, depicted in Table~\ref{tab:training_samples}, show that optimal performance is achieved with 200 samples per domain. We speculate that beyond this point, the number of training samples leads to the model over-fitting to domains in the training data and, therefore, less effective at zero-shot generalization. However, we anticipate that increasing the data size through increased diversity of datasets would lead to further improvements. 

\begin{table}[h!]
\centering
\resizebox{1.0\linewidth}{!}{
\begin{tabular}{l|ccccc|c}
\toprule
\#Data & Attr. & Hotel & Rest. & Taxi & Train & Avg.\\
\midrule
100 & 59.61 & 44.40 & 54.33 & 67.02 & 54.33 &55.94 \\ 
200 & 62.24 & \textbf{46.83} & \textbf{60.27} & \textbf{67.48} & \textbf{60.90} &\textbf{59.54}\\ 
300 & \textbf{69.19} & 43.68 & 57.06 & 64.98 & 57.60 &58.50\\ 
400 & 60.80 & 43.21 & 57.39 & 65.70 & 53.78 &56.18 \\ 
\bottomrule
\end{tabular}
}
\caption{\textsc{FnCTOD-LLaMA2-13B} with varying numbers of training data per domain (36 domains in total).}
\label{tab:training_samples}
\vspace{-5mm}
\end{table}


\section{Conclusion}
\vspace{-2mm}
We introduce a new approach to tackle the challenging task of zero-shot DST with LLMs, enabling them to handle both general conversations and task-oriented dialogues in diverse domains without the need for additional data collection. Our experimental results on MultiWOZ demonstrate that our approach not only delivers exceptional performance in advanced ChatGPT models (setting a new benchmark) but also across a range of moderately sized open-source LLMs. Furthermore, we demonstrate that we can fine-tune the open-source model \textsc{LLaMA-2-13B-Chat} using only 7,200 training samples from 36 diverse domains, resulting in \textsc{FnCTOD-LLaMA2-13B}, which achieves function calling, zero-shot DST performance comparable to ChatGPT.


\section{Limitations}
In this work, we propose a novel approach to solve zero-shot DST with LLMs. Our approach achieves outstanding performance with various LLMs, both modestly-sized open-source and advanced proprietary LLMs, setting the new state-of-the-art. However, it is important to recognize that the current accuracy may still not be good enough for the practical deployment of such zero-shot systems. We anticipate that with further advancements in the NLU and NLG capabilities of base LLMs, our approach could achieve even greater performance levels.
In addition, our approach can handle both the DST and response generation task in TOD. We evaluate DST with the well-established metric JGA with results suggesting the strong zero-shot DST performance of our approach. For the response evaluation, due to the current lack of a more realistic evaluation setting for response generation in TOD, we evaluated delexicalized responses as this is widely used in prior work. This setting and associated metrics have some known shortfalls in terms of being able to game-the-metrics with nonnatural responses as well as presenting a data mismatch with how LLMs are trained. In the era of LLMs, we advocate for the development of more realistic evaluation approaches for full-natural-language-response generation in TOD. Additionally, while this work concentrates DST and response generation, the two critical tasks in TOD, our approach can also be extended to include other complex tasks and handle various scenarios in dialogues as in the LangChain framework. We plan to explore these extensions in future research.



\bibliography{anthology,custom}
\bibliographystyle{acl_natbib}

\newpage
\appendix

\section{Appendix}
\label{sec:appendix}

\subsection{Evaluation Details}
\paragraph{Model and Inference Details}
We evaluated two versions of ChatGPT and six leading chat/instruction-tuned LLMs representing varying sizes and instruction-following and conversational capabilities. The six evaluated open-source models include: \textsc{Zephyr-7B-Beta}~\citep{zephyr} is an instruction-tuned version of Mistral-7B~\citep{mistral}, which is the leading model among its size on the AlpacaEval leaderboard~\cite{alpacaeval}. \textsc{Vicuna-7B-v1.5} and \textsc{Vicuna-13B-v1.5}~\citep{vicuna2023} are \textsc{LLaMA-2} models fine-tuned on user conversations with ChatGPT. \textsc{LLaMA2-7B-Chat} and \textsc{LLaMA2-13B-Chat} are chat-tuned versions of \textsc{LLaMA2} models with varying sizes~\citep{llama2}. \textsc{Baichuan2-13B-Chat} is also a \textsc{LLaMA2-13B} model further fine-tuned on extensive corpus~\cite{baichuan2}.
We utilized checkpoints available on Huggingface\footnote{\url{https://huggingface.co/models}}. The specific paths for these models are detailed in Table~\ref{tab:model_path}. 
For inference, the temperature was fixed as 0.3, top\_p as 0.2, and max\_tokens as 128.
For each test case, we conducted a single inference run. All inferences were executed on a cluster equipped with eight 48G NVIDIA RTX A6000 GPUs. 

\paragraph{End-to-end Evaluation Setup}
Our end-to-end evaluation of task-oriented dialogues follows well-established standards in the literature. 
The traditional TOD computational process contains multiple steps, of which this work focuses on two: (1) dialogue state tracking (DST) which tracks the mentioned slots in order to query a back-end knowledge source, e.g., database or API, and (2) response generation (NLG) wherein the model generates a response given the context and query results. Following the literature, our end-to-end evaluation includes the evaluation of DST with JGA (our main results) \emph{at the turn level}, and the evaluation of generated responses (NLG) using ``Success rate'' \emph{at the dialogue level}. The latter measures the percentage of dialogues in which the user's goals were fully met. We do not use or evaluate dialog acts, as they are not necessary for response generation.

The responses of the TOD system should provide specific information about a set of entities as requested by the user, like a restaurant or hotel. As multiple entities may satisfy the user's goal and it can be difficult to detect whether the response relates to a matching entity, delexicalized responses have been used in previous literature to allow for easier measurement and to decouple NLG errors from retrieval errors. In a delexicalized response, the mentioned slot values are noted as [value\_xxx]. For instance, a delexicalized response might be: ``It is a restaurant that serves [value\_food] food near the [value\_area]. The address is [value\_address] and the phone number is [value\_phone].'' This strategy facilitates easier automatic determination of whether the dialogue answers specific requests by the conversation's end. If both, all entities of the type expected are mentioned, and provision of all required information about those entities is also detected, then the dialogue is considered successful \cite{budzianowski-etal-2018-multiwoz}. This results in a binary ``Success'' metric per dialogue. Regardless of the user's input, the evaluation is on whether the entities and information about those entities provided to the user by the end of the conversation satisfies their goal. For example, given a user goal at the beginning of a dialogue to obtain the phone number of a restaurant serving French cuisine, the dialogue is marked as a success if by the end the system has mentioned at least one restaurant and provided a restaurant phone number. While JGA indicates whether the correct search for a restaurant serving French food was attempted.

\subsection{Training Details}

\begin{table}[h!]
\centering
\footnotesize
\resizebox{0.42\textwidth}{!}{
\begin{tabular}{l|l}
\toprule
\textbf{Hyperparameter} & \textbf{Values} \\ 
\hline
\hline
batch size & $8$\\
epochs & $1$ \\
learning rate&  $0.0003$ \\
learning rate scheduler&  cosine \\
weight decay&  $0.01$ \\
cutoff\_len& $4096$ \\
lora\_r & $16$ \\
lora\_alpha & $16$ \\
lora\_dropout & $0.05$ \\
lora\_target\_modules & q\_proj, v\_proj \\
\bottomrule
\end{tabular}
}
\caption{Hyperparameters for the model fine-tuning.}
       \label{tab:hyperparameter}
\end{table}

\paragraph{Training Data}
For constructing our fine-tuning dataset, we selected five high-quality, multi-turn TOD corpora, excluding MultiWOZ, as detailed in Table~\ref{tab:datasets}. Each dataset encompasses one or multiple domains. We excluded several domains with low-quality annotations, retaining a total of 36 domains. For our fine-tuning, we exclusively sampled data from the training sets of these datasets to constitute our training data.




\begin{table*}[tb]
    \vspace{-2mm}
    \small
    \centering
        \resizebox{0.9\linewidth}{!}{
        \begin{tabular}{l|l}
            \hline
             \textbf{Model} & \textbf{Model versioning/path} \\ \hline
            \hline
            GPT-3.5-Turbo &\texttt{gpt-3.5-turbo-1106}\\
            GPT-4 &\texttt{gpt-4-1106-preview}\\
            \hline
            Zephyr-7B-Beta & \url{https://huggingface.co/HuggingFaceH4/zephyr-7b-beta}\\
            Vicuna-7B-v1.5 & \url{https://huggingface.co/lmsys/vicuna-7b-v1.5}\\
            Vicuna-13B-v1.5 & \url{https://huggingface.co/lmsys/vicuna-13b-v1.5}\\
            Baichuan2-13B-Chat & \url{https://huggingface.co/baichuan-inc/Baichuan2-13B-Chat}\\
            LLaMA2-7B-Chat & \url{https://huggingface.co/meta-llama/Llama-2-7b-chat-hf}\\
            LLaMA2-13B-Chat & \url{https://huggingface.co/meta-llama/Llama-2-13b-chat-hf}\\
            \hline
        \end{tabular}
        }
    \caption{
        Evaluated LLMs in our experiments with their versions or Huggingface model paths.
    }
    \label{tab:model_path}
\end{table*}

\begin{table*}[tb]
\centering
\footnotesize
\resizebox{1.0\textwidth}{!}{
\begin{tabular}{lll}
\toprule
\textbf{Dataset}
& \multicolumn{1}{l}{\textbf{Domains}} & \textbf{\#Domains}\\ 
\cmidrule{1-3}
Schema-Guided~\cite{sgd} & RentalCars\_1, RentalCars\_2, Buses\_1, Buses\_2, Events\_1, Events\_2,   & 26\\ 
 & Services\_1, Services\_2, Services\_3, Media\_1, RideSharing\_1, RideSharing\_2, \\
& Travel\_1, Hotels\_1, Hotels\_2, Hotels\_3, Flights\_1, Flights\_2, Restaurants\_1,\\
 & Calendar\_1, Music\_1, Music\_2, Weather\_1, Movies\_1, Homes\_1, Banks\_1 \\
\cmidrule{1-3}
CamRest676~\cite{camres6761} & Restaurant & 1\\
\cmidrule{1-3}
MSR-E2E~\cite{mse2e} & Restaurant, Movie, Taxi & 3 \\
\cmidrule{1-3}
TaskMaster~\cite{taskmaster} & pizza\_ordering, movie\_ticket, auto\_repair,  uber\_lyft, coffee\_ordering & 5 \\
\cmidrule{1-3}
WOZ~\cite{woz} & Restaurant & 1\\
\bottomrule
\end{tabular}
}
\caption{Overview of the multi-turn TOD corpora utilized for fine-tuning, comprising a total of 36 diverse domains. This table details the datasets along with their specific domains and the number of domains included in each dataset.}\label{tab:datasets}
\end{table*}

\paragraph{Hyperparameters}
We fine-tuned the LLaMA-2-13b-Chat checkpoint from Hugginface.\footnote{\url{https://huggingface.co/meta-LLaMA/LLaMA-2-13b-chat-hf}}. We utilize Low Rank Approximation (LoRA)~\cite{lora} and limited our fine-tuning to the parameters in the \texttt{q\_proj} and \texttt{v\_proj} modules. Further details about the fine-tuning hyperparameters can be found in Table~\ref{tab:hyperparameter}. The fine-tuning was conducted on 4 A6000 48GB GPUs.

\begin{table*}[tb]
\centering
\resizebox{1.0\linewidth}{!}{
\begin{tabular}{lcccccccccccccc}
\toprule
\multirow{2}{*}{\textbf{Model}} & \multicolumn{2}{c}{Attraction} & \multicolumn{2}{c}{Hotel} & \multicolumn{2}{c}{Restaurant} & \multicolumn{2}{c}{Taxi} & \multicolumn{2}{c}{Train} & \multicolumn{2}{c}{\textbf{JGA}} & \multicolumn{2}{c}{\textbf{NLG}} \\
		\cmidrule(lr){2-3}
		\cmidrule(lr){4-5}
		\cmidrule(lr){6-7}
		\cmidrule(lr){8-9}
  		\cmidrule(lr){10-11}
		\cmidrule(lr){12-13}
            \cmidrule(lr){14-15}
& \textbf{JGA} & \textbf{F1} & \textbf{JGA} & \textbf{F1} & \textbf{JGA} & \textbf{F1} & \textbf{JGA} & \textbf{F1} & \textbf{JGA} & \textbf{F1} & \textbf{Average} & \textbf{Overall} & \textbf{Inform} & \textbf{Success}\\
\midrule

 

\rowcolor{mygreen1}
ChatGPT~\cite{hudevcek2023llms} & - & - & - & - & - & - & - & - & - & - & - & 27.00 & - & 44.00\\

\rowcolor{lightcyan}
\textsc{FnCTOD-LLaMA2-13B} & 62.43 & 85.55 & \textbf{46.49} & \textbf{84.92} & 61.51 & 89.36 & 69.11 & 81.13 & \textbf{61.77} & \textbf{90.73} & \textbf{60.26} & \textbf{37.94} & \textbf{85.10} & 44.50 \\

\rowcolor{lightyellow}
\textsc{Zephyr-7B-Beta} & 56.27 & 81.83 & 38.74 & 79.64 & \textbf{62.91} & \textbf{91.16} & \textbf{74.03} & \textbf{86.26} & 56.76 & 90.15 & 57.74 & 32.33 & 74.40 & \textbf{57.40} \\

\rowcolor{lightyellow}
\textsc{Vicuna-7B-v1.5} & 50.66 & 75.01 & 35.28 & 73.69 & 52.91 & 85.49 & 67.28 & 80.65 & 59.42 & 88.80 & 53.11 & 29.48 &66.70 &37.70 \\

\rowcolor{lightyellow}
\textsc{Vicuna-13B-v1.5} & 54.51 & 81.16 & 42.24 & 82.09 & 56.50 & 87.29 & 70.75 & 84.28 & 60.69 & 89.38 & 56.94 & 33.87 & 62.30 & 23.10 \\

\rowcolor{lightyellow}
\textsc{Baichuan2-13B-Chat} & 53.13 & 79.98 & 40.43 & 81.78 & 58.90 & 87.84 & 69.11 & 82.24 & 60.69 & 89.37 & 56.45 & 33.02 & 67.70 & 45.70 \\

\rowcolor{lightyellow}
\textsc{LLaMA2-7B-Chat} & 43.05 & 70.41 & 30.44 & 69.61 & 37.48 & 78.79 & 61.77 & 73.11 & 43.74 & 82.17 & 43.30 & 16.68 & 63.60 &24.90 \\

\rowcolor{lightyellow}
\textsc{LLaMA2-13B-Chat} & 49.95 & 76.91 & 29.53 & 67.58 & 48.64 & 81.44 & 64.72 & 68.90 & 53.64 & 85.02 & 49.30 & 25.83 & 64.20 & 27.70 \\

\bottomrule
\end{tabular}
}
\caption{\textbf{Detailed end-to-end evaluation results on MultiWOZ 2.2}, complementing Table~\ref{tab:nlg}. The best performances in each column are in \textbf{bold}.}
\label{tab:nlg_more}
\end{table*}

\subsection{Supplementary Results}

\begin{table*}[tb]
\centering
\resizebox{1.0\linewidth}{!}{
\begin{tabular}{lcccccccccccccc}
\toprule
\multirow{2}{*}{\textbf{Model}} & \multirow{2}{*}{\makecell[c]{\textbf{Oracle} \\ \textbf{Domain}}} & \multirow{2}{*}{\makecell[c]{\textbf{FS} \\ \textbf{Acc.}}} & \multicolumn{2}{c}{Attraction} & \multicolumn{2}{c}{Hotel} & \multicolumn{2}{c}{Restaurant} & \multicolumn{2}{c}{Taxi} & \multicolumn{2}{c}{Train} & \multicolumn{2}{c}{\textbf{JGA}}\\
		\cmidrule(lr){4-5}
		\cmidrule(lr){6-7}
		\cmidrule(lr){8-9}
  		\cmidrule(lr){10-11}
		\cmidrule(lr){12-13}
            \cmidrule(lr){14-15}
& & & \textbf{JGA} & \textbf{F1} & \textbf{JGA} & \textbf{F1} & \textbf{JGA} & \textbf{F1} & \textbf{JGA} & \textbf{F1} & \textbf{JGA} & \textbf{F1} & \textbf{Avg.} & \textbf{Overall} \\
\midrule

\rowcolor{lightcyan}
ChatGPT (GPT-4) & $\times$ & 88.62 & 58.77 & 81.84 & 45.15 & 85.07 & 63.18 & 91.06 & 76.39 & 87.73 & 69.48 & 90.16 & 62.59 & 38.71\\ 
\rowcolor{lightcyan}
ChatGPT (GPT-4) & $\checkmark$ & - & 65.61 & 87.42 & 49.08 & 87.07 & 64.54 & 90.99 & 77.97 & 88.53 & 72.12 & 93.02 & 65.86 & 44.70  \\ 

ChatGPT (GPT-3.5) & $\times$ & 95.54 & 67.15 & 87.20 & 37.56 & 82.86 & 60.12 & 90.21 & 74.43 & 86.90 & 67.29 & 92.48 & 61.31 & 38.56 \\
ChatGPT (GPT-3.5) & $\checkmark$ & - & 66.38 & 86.97 & 37.03 & 82.60 & 60.98 & 90.12 & 75.28 & 86.98 & 64.52 & 91.85 &60.84 & 38.32 \\ \hline

\rowcolor{lightcyan}
\textsc{FnCTOD-LLaMA2-13B} & $\times$ & 91.68 & 62.24 & 84.99 & 46.83 & 85.39 & 60.27 & 88.69 & 67.48 & 80.39 & 60.90 & 89.88 & 59.54 & 37.67 \\
\rowcolor{lightcyan}
\textsc{FnCTOD-LLaMA2-13B} & $\checkmark$ & - & 62.88 & 85.91 & 47.55 & 85.33 & 59.67 & 88.61 & 71.02 & 82.36 &61.61 & 90.25 & 60.55 & 38.70\\

\textsc{Zephyr-7B-Beta} & $\times$ & 92.77 & 56.50 & 81.97 & 38.43 & 79.52 & 63.18 & 91.19 & 74.10 & 86.56 & 56.20 & 90.00 & 57.68 & 32.11\\
\textsc{Zephyr-7B-Beta} & $\checkmark$ & - & 57.78 & 83.05 & 42.09 & 83.91 & 62.46 & 90.39 & 78.23 & 88.18 & 56.20 & 89.93 & 59.35 & 34.40 \\ 

\rowcolor{lightcyan}
\textsc{Vicuna-7B-v1.5} & $\times$ & 94.75 & 50.66 & 74.93 & 35.00 & 73.66 & 52.76 & 85.25 & 67.02 & 80.36 & 59.66 & 89.05 & 53.02 & 29.45 \\
\rowcolor{lightcyan}
\textsc{Vicuna-7B-v1.5} & $\checkmark$ & - & 50.59 & 74.80 & 36.00 & 74.38 & 52.58 & 84.75 & 69.11 & 81.81 & 59.13 & 88.39 & 53.48 & 30.06 \\

\textsc{Vicuna-13B-v1.5} & $\times$ & 91.82 & 54.25 & 80.99 & 38.43 & 79.96 & 56.44 & 87.26 & 69.11 & 83.37 & 58.82 & 89.26 & 55.41 & 31.84 \\
\textsc{Vicuna-13B-v1.5} & $\checkmark$ & - & 54.80 & 81.28 & 42.24 & 81.99 & 54.93 & 86.55 & 71.41 & 84.92 & 60.69 & 88.96 & 56.81 & 34.20 \\ 

\rowcolor{lightcyan}
\textsc{Baichuan2-13B-Chat} & $\times$ & 92.50 & 53.67 & 79.57 & 40.15 & 81.36 & 59.02 & 87.82 & 69.31 & 81.95 & 60.67 & 89.45 & 56.56 & 33.21 \\
\rowcolor{lightcyan}
\textsc{Baichuan2-13B-Chat} & $\checkmark$ & - & 56.43 & 80.62 & 40.34 & 82.55 & 59.88 & 88.95 & 69.84 & 82.64 & 61.35 & 89.23 & 56.97 & 34.94 \\

\textsc{LLaMA2-7B-Chat} & $\times$ & 91.90 & 42.64 & 70.18 & 30.47 & 69.37 & 37.60 & 78.63 & 63.20 & 73.80 & 44.17 & 82.18 & 43.44 & 16.78 \\
\textsc{LLaMA2-7B-Chat} & $\checkmark$ & - & 45.36 & 72.92 & 30.78 & 69.89 & 38.43 & 79.29 & 65.05 & 75.85 & 43.40 & 81.65 & 44.60 & 18.75 \\

\rowcolor{lightcyan}
\textsc{LLaMA2-13B-Chat} & $\times$ & 89.34 & 49.76 & 76.80 & 29.50 & 67.60 & 48.87 & 81.33 & 64.66 & 68.97 & 53.59 & 85.09 & 49.28 & 25.68 \\
\rowcolor{lightcyan}
\textsc{LLaMA2-13B-Chat} & $\checkmark$ & - & 50.40 & 77.59 & 30.28 & 68.91 & 49.08 & 81.09 & 65.97 & 69.70 & 54.80 & 84.89 & 50.11 &  29.56 \\


\bottomrule
\end{tabular}
}
\caption{Detailed results regarding model performances with oracle and predicted domains/functions, complementing Table~\ref{tab:dp}. FS Acc. indicates the function selection accuracy, and the JGA here indicates overall JGA.}
\label{tab:dp_more}
\end{table*}

In this section, we present the detailed figures from the experimental results depicted in Table~\ref{tab:nlg}, Table~\ref{tab:dp}, Figure~\ref{fig:fs_ablation}, and Figure~\ref{fig:nshot}. Specifically, additional results from the end-to-end evaluation on MultiWOZ 2.2 are presented in Table~\ref{tab:nlg_more} (complementing Table~\ref{tab:nlg}).
Further results from the ablation studies with predicted and oracle functions are presented in Table~\ref{tab:dp_more} (complementing Table~\ref{tab:dp}).

\subsection{Details of Prompt Formatting}

\paragraph{Conversation Context}
To format conversations within the prompt, we adopted the specific chat format for each LLM evaluated, as used in their respective instruction tuning steps.\footnote{\url{https://github.com/lm-sys/FastChat}}

\paragraph{System prompt}
In our evaluation, we utilized the following system prompt template:
\begin{tcolorbox}[title = {System prompt}]
\small
You are a task-oriented assistant. You can use the given functions to fetch further data to help the users. \\
\\
<FUNCTIONS> \\
\textcolor{blue}{\{Function Specifications\}}\\
</FUNCTIONS>\\
\\
<EXAMPLES> \\
\textcolor{blue}{\{Example Conversations\}}\\
</EXAMPLES>
\end{tcolorbox}
The parts surrounded in brackets and highlighted in blue serve as placeholders and are replaced with specific function specifications and example conversations related to that function/domain. The example part is only employed for few-shot prompting with the models not fine-tuned for function-calling.

\begin{figure*}[!t]
\begin{center}
\includegraphics[width=1\linewidth]{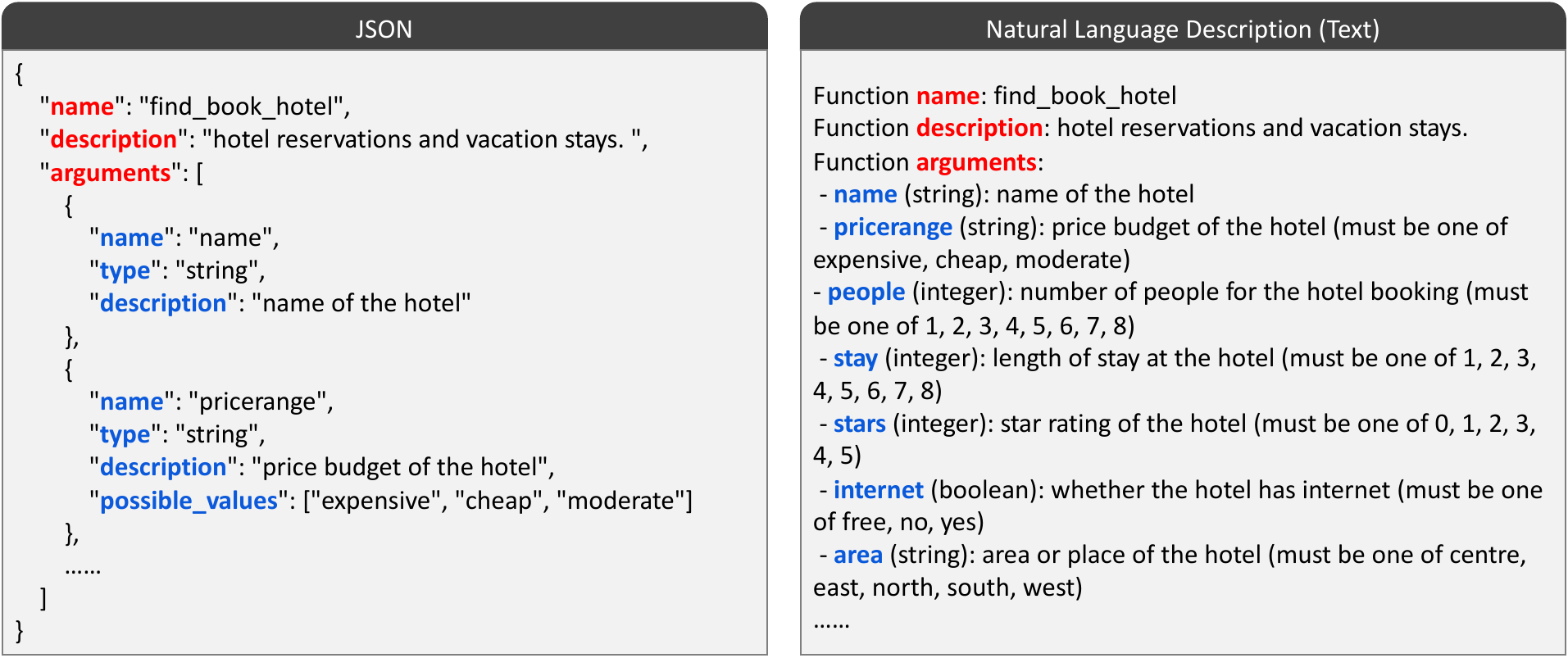}
\end{center}
   \caption{The JSON format (left) and its corresponding natural language description (right) utilized in our evaluation. We take the hotel domain as an example.
   }\label{fig:type}
\vspace{-3mm}
\end{figure*}

\paragraph{Function Specifications}
For the function specification within the system prompt section of the prompt, we adhere to ChatGPT's format. To enhance model comprehension, we also experimented with translating the JSON format into a natural language description to include in the system prompt. An example illustrating both the JSON format and its corresponding natural language description for a specific domain is depicted in Figure~\ref{fig:type}.

\begin{figure*}[!t]
\begin{center}
\includegraphics[width=1\linewidth]{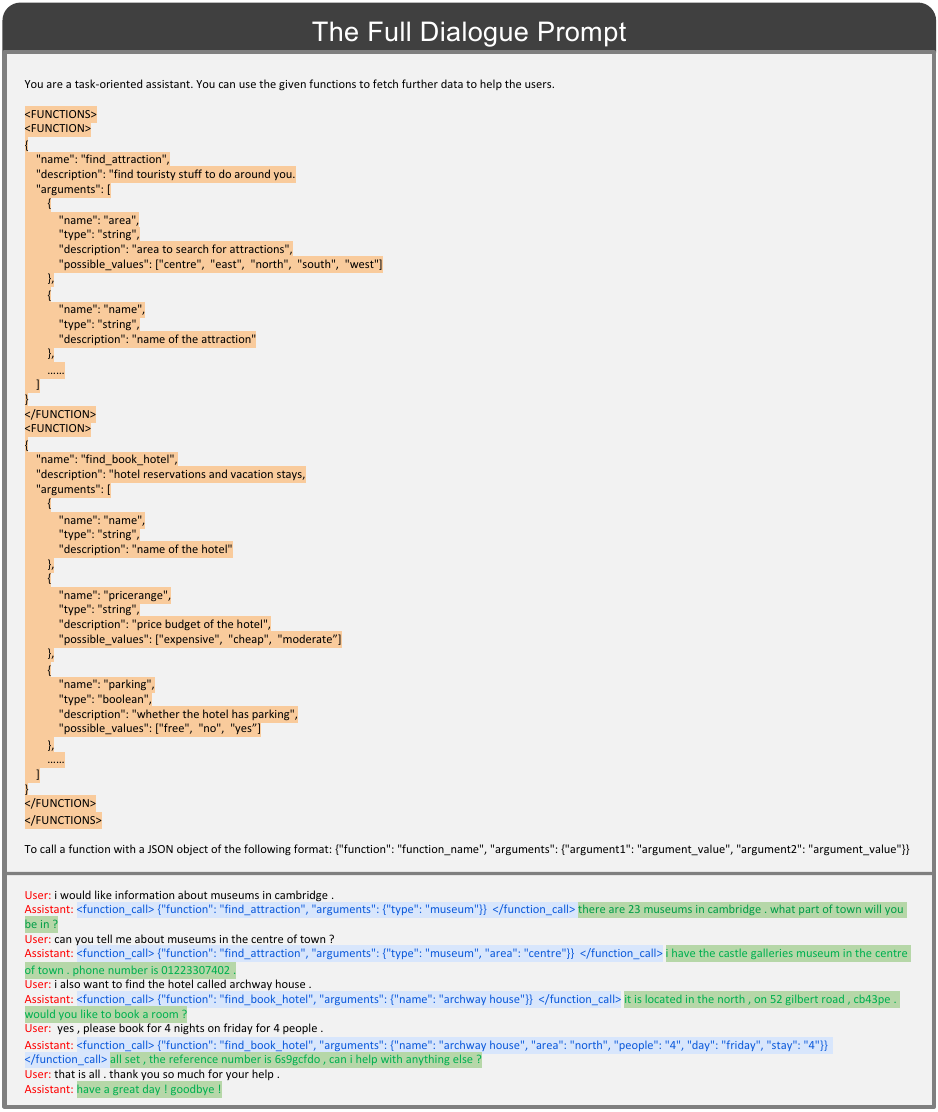}
\end{center}
   \caption{The complete dialogue format employed for model fine-tuning with no demonstration examples. Key components such as the \colorbox{myyellow1}{function specifications} in the system prompt, the \colorbox{myblue1}{function call}, and the \colorbox{mygreen1}{response} in the assistant's output are highlighted for clarity. It's important to note that for easier comprehension, we've adopted a more human-readable dialogue format, and have simplified some function arguments for clearer visualization. In practice, we use the specific chat format tailored to each evaluated model, and the LLaMA2-Chat format is applied for fine-tuning the LLaMA2-Chat model.
   }\label{fig:full_prompt}
\vspace{-3mm}
\end{figure*}

\paragraph{Full Prompt}
Combining all components, an example of the full dialogue prompt is displayed in Figure~\ref{fig:full_prompt}. For clearer illustration, we adopt a more human-readable dialogue format not including the special tokens used in model-specific dialogue formats.

\end{document}